\documentclass[10pt,twocolumn]{IEEEtran}
\usepackage{hyperref}
\usepackage{url}

\usepackage{algorithm}
\usepackage{algorithmic,color}
\usepackage{amsfonts}
\usepackage{amsmath}
\usepackage{amssymb}
\usepackage{amsthm}
\usepackage{float}
\usepackage{framed}
\usepackage{wrapfig}

\newcommand{\barr}{\left[ \begin{array} }
\newcommand{\earr}{ \end{array} \right] }

\newcommand{\ars}[1]{\left[ \begin{array}{#1}}
\newcommand{\are}{\end{array} \right] }
\newcommand{\oars}[1]{\begin{array}{#1}}
\newcommand{\oare}{\end{array}}

\newcommand{\eqs}{\begin{eqnarray}}
\newcommand{\eqe}{\end{eqnarray}}
\newcommand{\eqsn}{\begin{eqnarray*}}
\newcommand{\eqen}{\end{eqnarray*}}

\newcommand{\ens}{\begin{enumerate}}
\newcommand{\ene}{\end{enumerate}}

\newcommand{\its}{\begin{itemize}}
\newcommand{\ite}{\end{itemize}}

\newcommand{\des}{\begin{description}}
\newcommand{\dee}{\end{description}}
\newcommand{\Tr}{\mathrm{Tr}}
\newcommand{\cH}{{\cal H}}
\newcommand{\hcH}{\hat{{\cal H}}}

\usepackage{mathrsfs}
\DeclareSymbolFontAlphabet{\mathrsfs}{rsfs}
\usepackage[mathscr]{eucal}

\usepackage{amsbsy}

\usepackage{bm}

\usepackage{paralist}

\usepackage{xspace}

\newcommand{\Sec}[1]{\hyperref[sec:#1]{\S\ref*{sec:#1}}} 
\newcommand{\Eqn}[1]{\hyperref[eq:#1]{(\ref*{eq:#1})}} 
\newcommand{\Fig}[1]{\hyperref[fig:#1]{Figure~\ref*{fig:#1}}} 
\newcommand{\Tab}[1]{\hyperref[tab:#1]{Table~\ref*{tab:#1}}} 
\newcommand{\Thm}[1]{\hyperref[thm:#1]{Theorem~\ref*{thm:#1}}} 
\newcommand{\Lem}[1]{\hyperref[lem:#1]{Lemma~\ref*{lem:#1}}} 
\newcommand{\Prop}[1]{\hyperref[prop:#1]{Property~\ref*{prop:#1}}} 
\newcommand{\Cor}[1]{\hyperref[cor:#1]{Corollary~\ref*{cor:#1}}} 
\newcommand{\Def}[1]{\hyperref[def:#1]{Definition~\ref*{def:#1}}} 
\newcommand{\Alg}[1]{\hyperref[alg:#1]{Algorithm~\ref*{alg:#1}}} 
\newcommand{\Ex}[1]{\hyperref[ex:#1]{Example~\ref*{ex:#1}}} 

\newcommand{\Real}{\mathbb{R}}
\newcommand{\Cplx}{\mathbb{C}}

\newcommand{\mc}[1]{\mathcal{#1}}

\newcommand{\mcr}[1]{\mathrsfs{#1}}

\newcommand{\V}[1]{\bm{\mathbf{#1}}} 

\newcommand{\M}[1]{{\bm{\mathbf{\MakeUppercase{#1}}}}} 

\newcommand{\T}[1]{\boldsymbol{\mathscr{\MakeUppercase{#1}}}} 

\newcommand{\fft}{ \mbox{\tt fft} }
\newcommand{\ifft}{ \mbox{\tt ifft} }
\newcommand{\blkd}{\mbox{\tt blkdiag}}
\newcommand{\rshpT}{\mbox{\tt reshapeT}}

\newtheorem{theorem}{Theorem}[section]
\newtheorem{corollary}{Corollary}[section]
\newtheorem{lemma}{Lemma}[section]
\theoremstyle{definition}
\newtheorem{definition}{Definition}[section]

\usepackage{graphicx} 
\usepackage{subfigure}

\ifCLASSINFOpdf
\else
\fi
\begin{document}
%
\title{An Algorithm for Online Tensor Prediction}
%
%
%

\author{John~Pothier,
        Josh Girson,
        and~Shuchin~Aeron,~\IEEEmembership{Member,~IEEE}
\thanks{John Pothier is with Microsoft Seattle. This work was done in part when John was an undergraduate summer researcher at Tufts during 2013-2014.}
\thanks{Josh Girson and Shuchin Aeron are with the Dept. of ECE at Tufts University. The research is supported by NSF CCF: 1319653 and NSF Research Experiences for Undergraduates (REU). The corresponding author can be reached at shuchin@ece.tufts.edu}
\thanks{}}

\maketitle

\begin{abstract} 
We present a new method for online prediction and learning of tensors ($N$-way arrays $N >2$) from sequential measurements. We focus on the specific case of 3-D tensors and exploit a recently developed framework of structured tensor decompositions proposed in \cite{KilmerMartin2010}. In this framework it is possible to treat 3-D tensors as linear operators and appropriately generalize notions of rank and positive definiteness to tensors in a natural way. Using these notions we propose a generalization of the matrix exponentiated gradient descent algorithm \cite{Tsuda2005wc} to a tensor exponentiated gradient descent algorithm using an extension of the notion of von-Neumann divergence to tensors. Then following a similar construction as in \cite{Hazan2012:un}, we exploit this algorithm to propose an online algorithm for learning and prediction of tensors with provable regret guarantees. Simulations results are presented on semi-synthetic data sets of ratings evolving in time under local influence over a social network. The result indicate superior performance compared to other (online) convex tensor completion methods. 
\end{abstract} 


%
\IEEEpeerreviewmaketitle

\section{Introduction}

The problem addressed by this paper is online prediction (completion) of 3-D arrays  $\T{M} \in  \Real^{n_1 \times n_2 \times n_3}$, also referred to as tensors\footnote{\textbf{\textcolor{blue}{Strictly speaking a tensor is a multilinear functional, mapping a collection of vectors to scalars and is linear in each argument separately. For finite dimensional vector spaces a tensor can be represented using a multidimensional array and hence the terminology.}}}. On each round $t$, the predictor (learner) receives a triplet of indices $(i_t,j_t,k_t)$ and predicts the value of $\T{M}(i_t,j_t,k_t)$. The learner then suffers a loss according to a convex loss function $l_t$, which is also selected \emph{adversarially} from a class of convex functions with bounded Lipschitz continuity. As is normally done in sequential estimation and learning \cite{Cesa_Book, ShalevShwartz:2012dz}, the goal is to minimize long term \emph{regret}, i.e. the loss compared to the best possible policy in hindsight,  over some class of predictors  (we make this precise in Section \ref{sec:prob_setup}).

Motivated by the success of low-rank heuristic for such problems for the case of 2-D arrays \cite{Koren10,ShamirS11,KakadeST12,Recht:2010:GMS:1958515.1958520}, to this end we will chose the comparator class based on the assumption that the best estimator belongs to a tensor with low \emph{tensor-rank}. In this context we exploit a recently proposed tensor factorization strategy proposed in \cite{Kilmer_SIAM13}. In this framework, the low-rank nature of a 3-rd order tensor is captured through a matrix like Singular Value Decomposition (SVD), namely  tensor-SVD (t-SVD). Similar to the case of algorithms used for matrix completion assuming that the data is sampled from a low rank matrix, recently similar methods based on the t-SVD have found success in tensor completion from missing entries for video (3D and 4D) \cite{zhang2014novel} and seismic (5D) data \cite{ely20135d}, and we are motivated by the results reported therein. However unlike the methods considered in these papers that assume a batch setting, in this paper we assume that the data is provided in a sequential or streaming manner and the goal is to minimize the long term (cumulative) prediction error. 


There are other approaches for tensor prediction in the batch and adaptive sampling situation using other types of tensor factorizations such as Canonincal-Parafac (CP) and Higher Order Singular Value Decomposition (HOSVD),\cite{Kolda09tensordecompositions,Wolfgang_Book12}. In contrast our work considers tensor prediction in an online and non-adaptive setting with performance guarantees. To the best of our knowledge the problem of non-adaptive online learning and prediction of tensors (in particular multidimensional data) has not been \emph{explicitly} considered so far. In order to put our contributions in perspective we begin by a survey of current frameworks used for modeling and prediction of tensor data. 

\subsection{Relation to existing work}
Existing work on tensor completion from limited measurements rely upon treating a tensor as an element of outer product of finite dimensional vector spaces \cite{Wolfgang_Book12}. Within this multilinear algebraic framework, tensor completion strategies under several rank-revealing factorizations namely Canonical/Parafac (CP) and Tucker \cite{SIREV} have been proposed, see \cite{Oh_NIPS14} for methods based on special cases (namely symmetric tensors) of CP decomposition and \cite{bartv_14} for methods based on Tucker and Hierarchical-Tucker decompositions. These methods essentially exploit the low rank matrix structure from various un-foldings and reshaping of the tensor. Put another way these methods assume that when a tensor is seen as an element of outer product of vector spaces, each vector space has low dimension. This fact is also exploited in a number of methods, which essentially work by deriving novel norms serving as a low rank convex surrogate, on the set of matrices obtained by mode unfoldings of a tensor   \cite{TomiokaSHK11}. Adaptive (non-adversarial) sampling and recovery methods have also been proposed \cite{ASingh_13}, which are again based on adaptively learning the vector spaces spanned by the tensor fibers. 

In contrast to these multilinear algebraic approaches our approach is linear algebraic and is based on the group theoretic approach of \cite{Kilmer_SIAM13,zhang2014novel}. At a high-level this approach essentially rests on unraveling the complexity of the multidimensional structure by constructing group-rings along the tensor fibers, \cite{Navasca_2010}. In this framework a 3-D tensor can be treated as a linear operator acting on the vector space over these group-rings. A rank revealing factorization of this operator then captures the complexity of the multidimensional data, which in turn is useful for prediction. In this paper we will restrict ourselves to cyclic groups, which can capture periodic patterns in the data.

\subsection{Organization of the paper}
We begin by noting necessary background material and preliminaries in Section \ref{sec:back}. In Section \ref{sec:psd} we derive notions of von Neumann entropy and divergence for tensors. Then in Section \ref{sec:prob_setup} we state the problem, outline the main results and derive Online Tensor Exponentiated Gradient (OTEG) descent algorithm. Simulation results on synthetic data sets are presented in Section \ref{sec:sim}.

\subsection{Notation}

Matrices will be denoted by upper case bold letters $\M{X}$, vectors by lower case boldface letters $\V{x}$ and 3-D arrays or tensors will be denoted by $\T{X}$. Throughout we will use the following notation for denoting the elements, fibers and slices for the tensors and matrices - for a tensor $\T{X}^{(i)}$ will denote the $i$-th \emph{frontal} slice of $\T{X}$, and $\T{X}_{ij}$ will denote a tensor fiber (or \emph{tube}) into the board. We will also use the following convention for denoting the tensor fibers - $\T{X}(:,:,k)$ denotes the $k$-th frontal face, $\T{X}(:,j,:)$ denotes the $j$-th lateral slice and $\T{X}(i,:,:,)$ denotes the $i$-th horizontal slice. Similarly $\M{X}(:,i)$ denotes the $i$-th column of the matrix and so on. For any third order tensor $\T{X}$, $\widehat{\T{X}}$ denotes the 3-D tensor of the same size obtained by taking the Fourier transform along the third dimension (also c.f. Algorithm \ref{alg:tSVD} in Section \ref{sec:tSVD}).

\section{Linear algebra for 3-D tensors }
\label{sec:back}
We will now briefly review the linear algebraic concepts first developed in \cite{KilmerMartin2010} shown to be useful in a variety of applications \cite{Hao_SIAM2013,zhang2014novel,ely20135d}. 

 
\subsection{t-product: Tensor as a linear operator}
\label{sec:tprod}
\begin{figure}[htbp]
       \vspace{-3mm}
 	\centering
 	\includegraphics[height=1.5in]{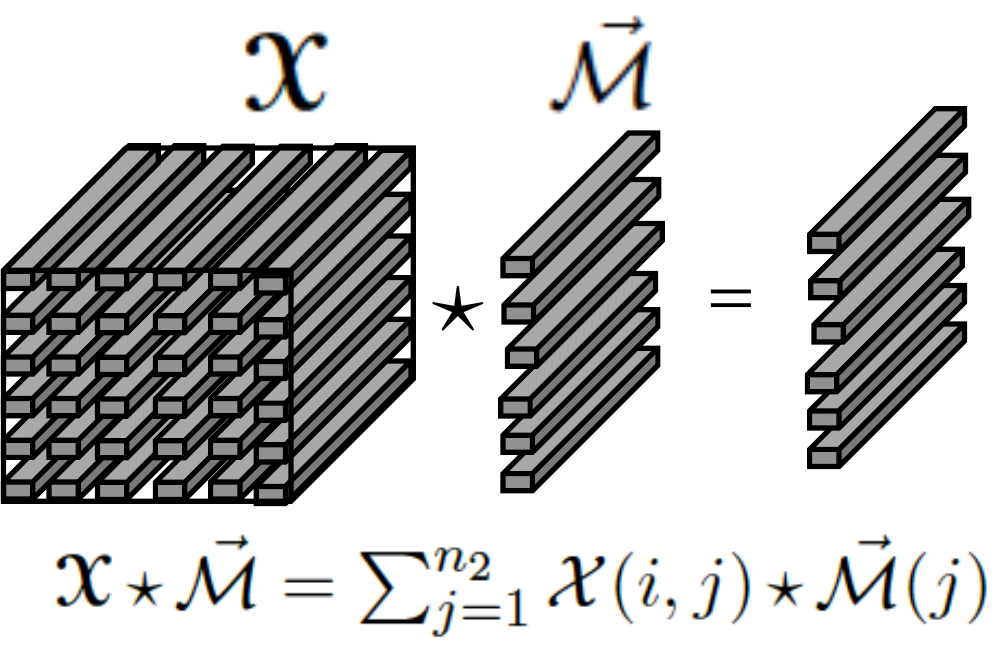}
 	\caption{\label{fig:t_prod} {3-D tensors as operators on oriented matrices.}}
 \end{figure} 

In the framework proposed in \cite{KilmerMartin2010} a 3-D array is defined as a linear operator using the t-product defining the multiplication action. There are several ways to define the t-product, and we take the development directly from \cite{zhang2014novel}. We begin by viewing a 3-D tensor $\T{X} \in \Real^{n_1 \times n_2 \times n_3}$ as an $n_1 \times n_2$ matrix (say) $\mathcal{X}$ of tubes (vectors oriented into the board), whose $i,j$-th entry $\mathcal{X}(i,j) = \T{X}(i,j,:)$. Similarly one can consider a $n_1 \times 1\times n_3$ tensor as a vector of tubes. Such tensors are referred to as oriented matrices, \cite{KilmerMartin2010} and are denoted by $\vec{\mc{M}}$. 
 
 Now in order to define the 3-D tensor as a linear operator on the set of oriented matrices $\vec{\mc{M}}$ \cite{Braman2010}, one defines a multiplication operation between two tubes $\vec{\V{v}} \in \Real^{1\times 1 \times n_3}$ and $\vec{\V{u}}  \in \Real^{1\times 1 \times n_3}$ resulting in another tube of same length. Specifically   this multiplication operation is given by circular convolution denoted by $\star$. Under this construction, the operation of a tensor $\T{X}$ on $\vec{\mc{M}} \in \Real^{n_2 \times 1 \times n_3}$ is another oriented matrix of size $n_1 \times 1 \times  n_3$ whose $i$-th tubal element given by, $ \T{X} \star \vec{\mc{M}} = \sum_{j=1}^{n_2} \mc{X}(i,j) \star \vec{\mc{M}}(j)$ as illustrated in Figure~\ref{fig:t_prod}. Similarly one can extend this definition to define the multiplication of two tensors $ \T{X}$ and $\T{Y}$ of sizes $n_1 \times n_2 \times n_3$ and $n_2 \times k \times n_3$ respectively, resulting in a tensor $\T{C} = \T{X} \star \T{Y}$ of size $n_1 \times k \times n_3$. This product between two tensors is referred to as the t-product.

\subsection{t-SVD}
\label{sec:tSVD}
\begin{figure} 
 	\centering
  \includegraphics[height= 2.25  in]{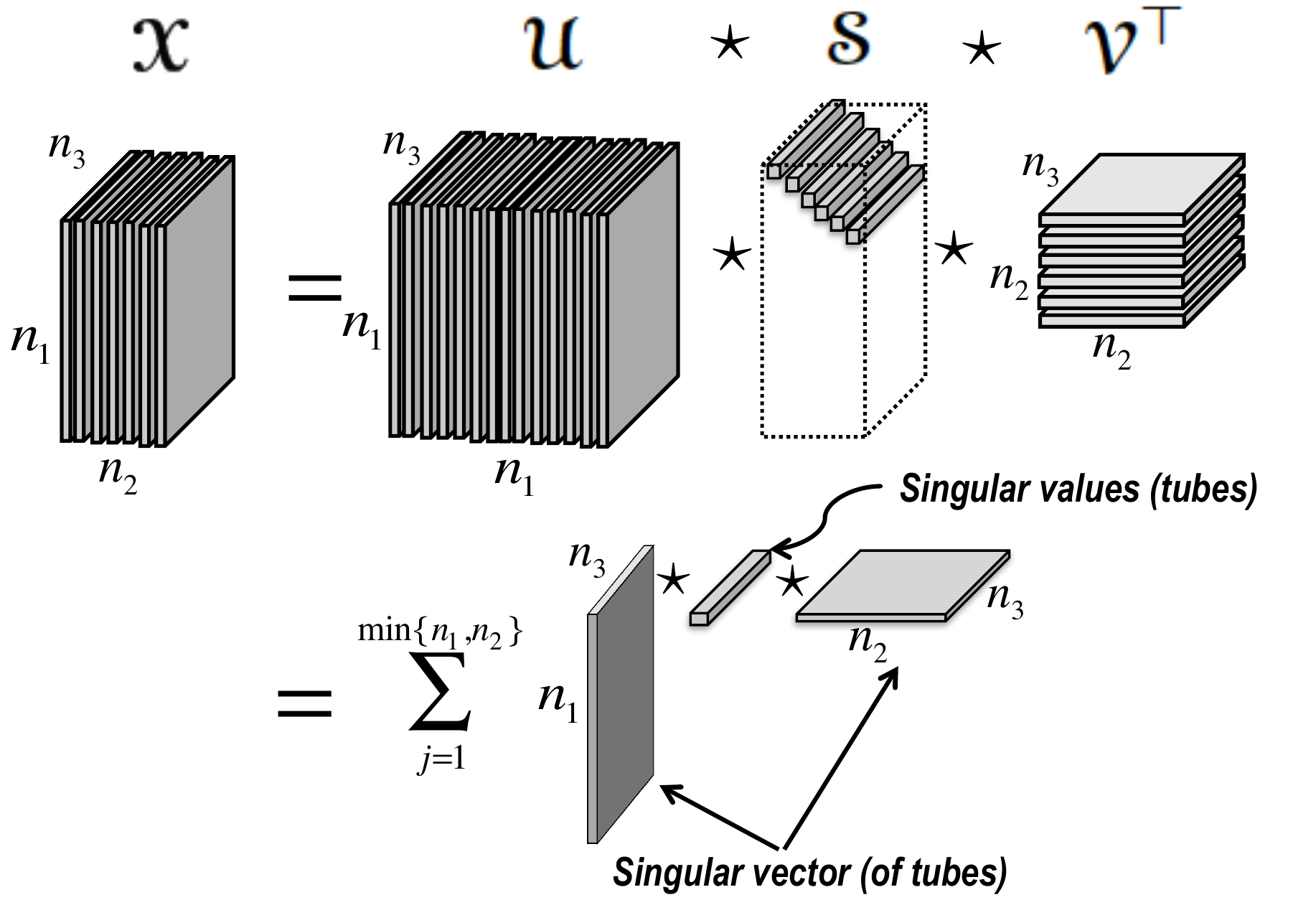}
  \caption{ {t-SVD under the t-product} }
\label{fig:tSVD}
\end{figure}

Under the above construction viewing a 3-D tensor as a linear operator over the set of oriented matrices, one can compute a tensor-Singular Value Decomposition (t-SVD) as shown in Figure~\ref{fig:tSVD}. Since  $\star$ is given by the circular convolution the t-SVD can be computed using the Fast Fourier Transform ($\fft$) using Algorithm \ref{alg:tSVD} \cite{KilmerMartin2010}. 

The component tensors $\T{U}$ and $\T{V}$ obey the orthogonality conditions $\T{U}^\top\star \T{U} = \T{I}$, $\T{V}^\top \star \T{V} = \T{I}$ with the following definitions for tensor transpose $(\cdot)^{\top}$ and  and identity tensor $\T{I}$ (of appropriate dimensions).
\begin{definition}
\emph{\textbf{Tensor Transpose}}.  Let $\T{X}$ be a tensor of size $n_1 \times n_2 \times n_3$, then $\T{X}^\top$ is the $n_2 \times n_1 \times n_3$ tensor obtained by transposing each of the frontal slices and then reversing the order of transposed frontal slices $2$ through $n_3$.
\end{definition}
\begin{definition} \emph{\textbf{Identity Tensor}}. The identity tensor $\T{I} \in \mathbb{R}^{n \times n \times n_3}$ is a tensor whose first frontal slice is the $n \times n$ identity matrix and  all other frontal slices are zero.
\end{definition}

 \begin{figure}[h]
 	\centering
    \begin{minipage}{0.5\textwidth}
\begin{algorithm}[H]
  \caption{tSVD}
  \begin{algorithmic}
   \label{alg:tSVD}
  \STATE \textbf{Input: } $\T{X} \in \mathbb{R}^{n_1 \times n_2  \times n_3}$
  \STATE Take Fourier transform along the 3 dimension
  	\STATE ${\widehat{\T{X}}} \leftarrow \fft(\T{X},[\hspace{1mm}],3)$;
  \FOR{$i = 1 \hspace{2mm} \rm{to} \hspace{2mm} n_3$}
  	\STATE $ [\hat{\M{U}}, \hat{\M{S}}, \hat{\M{V}}] = {\tt SVD} (\widehat{\T{X}}^{(i)})$
  	\STATE $ {\widehat{\T{U}}}^{(i)} = \hat{\M{U}};  {\widehat{\T{S}}}^{(i)} = \hat{\M{S}}$; $ \widehat{\T{V}}^{(i)}= \hat{\M{V}}; $
  \ENDFOR
  
  \STATE Take inverse Fast Fourier Transform ${\tt ifft}$ along the 3 dimension for each of the component tensors
  \STATE $\T{U} \leftarrow \ifft(\widehat{\T{U}},[\hspace{1mm}],3);$ $\T{S} \leftarrow \ifft(\widehat{\T{S}},[\hspace{1mm}],3);$ 
	\STATE $\T{V} \leftarrow \ifft(\widehat{\T{V}},[\hspace{1mm}],3)$;
  \end{algorithmic}
\end{algorithm}
\end{minipage}
\end{figure}

\subsubsection{Alegbraic Complexity measures from t-SVD}
Under the t-SVD, it is clear that \cite{Kilmer_SIAM13} if the number of non-zero singular tubes in $\T{S}$ is $r$, there exist a set of oriented matrices $\T{X}(:,j^\prime,:), j^\prime \in J: |J| = r$ such that each $\T{X}(:,i,:)$ can be written as $$\T{X}(:,j,:) = \sum_{j^\prime \in J} \T{X}(:,j^\prime,:) \star \vec{\V{\ell}}_{j\prime}^{j}.$$ 
From t-SVD one can readily extract several notions of complexity of the data in terms of ``rank". The notion of \emph{multi-rank} was proposed in \cite{Kilmer_SIAM13} using the Fourier Domain representation of t-SVD as the vector of ranks of the slices $\widehat{\T{X}}(:,:,i), i = 1,2,...,n_3$. The $\ell_1$ norm of the multi-rank can be taken to be a measure of the complexity of the data. On the other hand, similar to matrix completion, where the nuclear norm is used as a useful convex surrogate to low rank, one employs a similar measure form t-SVD known as Tensor Nuclear Norm (TNN) \cite{zhang2014novel}. TNN, denote by denoted $\|\T{X}\|_{TNN}$ is the sum of nuclear norms of the slices $\widehat{\T{X}}({:,:,i})$. In this paper we will derive complexity measures which are related to TNN and use them to define the class of predictors against which, we will find bounds on the regret.

\section{Positive-Definite~Tensors and von~Neumann~Entropy}
\label{sec:psd}

Based on the t-SVD we define the notion of positive definite tensors. 

\begin{definition}
\textbf{Positive Definite Tensor}: A tensor is positive definite under the t-product if each frontal slice $\widehat{\T{X}}^{(i)}$ in the transformed domain  is positive definite. 
\end{definition}
Similar definition applies to a symmetric positive definite tensors. In the following we will denote by $ \mcr{S}^{N \times N \times d}_{++} $ the set of all \emph{symmetric positive definite tensors}. 
 
\begin{definition} [Trace of a Tensor] The trace of the tensor $\T{X} \in \Real^{n_1 \times n_2 \times n_3}$ is defined as the trace of $\blkd(\widehat{\T{X}})$ where  $\blkd(\widehat{\T{X}})$ is a block diagonal matrix whose diagonal blocks are given by $\widehat{\T{X}}^{(i)}$. 
\end{definition}
Let $\rshpT(\blkd(\widehat{\T{X}}))$ denote the reshaping of $\blkd(\widehat{\T{X}})$ back to the tensor $\widehat{\T{X}}$ and $\T{X}^k = \underset{k\,\, \mbox{times}}{\underbrace{\T{X}\star \T{X} \star \hdots \star \T{X}}}$ for a positive integer $k$ and and $\T{X}^0=\T{I}$.

\begin{definition}
\label{def:exp}
Let $\T{X} \in \mcr{S}^{N \times N \times d}_{++} $. Then, under the t-product, we define the tensor exponential as
$
\exp (\T{X}) \triangleq \sum_{k=0}^\infty \frac{1}{k!}\T{X}^k
$.
\end{definition}
By a straightforward calculation it can be shown that $$\exp \T{X} = \ifft\left(\rshpT\left(\exp \left(\blkd(\widehat{\T{X}})\right)\right),[\,],3\right),$$ where the matrix exponential is defined in the usual way.
\begin{definition} [Logarithm of a Tensor] For $\T{X} \in \mcr{S}^{N \times N \times d}_{++} $, in line with the definition of tensor exponential, we define the logarithm of a tensor $\T{X}$ as $$
\log \T{X} \triangleq \ifft \left(\rshpT\left(\log \left(\blkd (\widehat{\T{X}})\right)\right) ,[\,], 3\right),$$
where the matrix logarithm is defined in the usual way.
\end{definition}


\subsection{Von-Neumann Entropy for Tensors}
We begin by extending the notion of Von-Neumann entropy for PD symmetric matrices \cite{Dhillon_SIAM2007} to PD tensors via the following.
\begin{definition} The von-Neumann entropy of a tensor $\in \mcr{S}^{N \times N \times d}_{++}$ is defined as 
\begin{align*}
& {\cal {H}}(\T{W}) \\ 
& \triangleq  \mathrm{Tr}\left(\blkd(\widehat{\T{W}}) \log(\blkd(\widehat{\T{W}}))- \blkd(\widehat{\T{W}})\right) \\
                  & = \sum_{k=1}^d \mathrm{Tr}\left(\widehat{\T{W}}^{(k)}\log(\widehat{\T{W}}^{(k)})-\widehat{\T{W}}^{(k)}\right)\\
                  & \triangleq \hat{\cH}(\widehat{\T{W}})
\end{align*}
\end{definition}
Note that by definition of the tensor trace, we can write, ${\cal {H}}(\T{W})  = \mathrm{Tr}(\T{W}\star \log \T{W} - \T{W})$. Let us define an inner product on 
the space of real tensors via the t-product. 

\begin{definition}
[Inner product of two tensors] The inner product between two $n_1\times n_2 \times n_3$ tensors $\T{X}, \T{Y} $ is defined as 
\begin{align*}
\langle \T{X}, \T{Y} \rangle = \Tr(\T{X} \star \T{Y}^\top) = \Tr(\blkd(\widehat{\T{X}})\blkd(\widehat{\T{Y}})^\dagger)\,\, ,
\end{align*}
\end{definition}
where $\dagger$ denotes Hermitian transpose.

We now derive the von-Neumann divergence $\Delta_H(\T{W}',\T{W})$ between tensors $\T{W},\T{W}' \in \mcr{S}^{N \times N \times d}_{++}$. Note that under the t-product we have,
\begin{align*}
& \Delta_H(\T{W}',\T{W}) \nonumber \\
& = {\cal H}(\T{W}') - {\cal H}(\T{W}) - \Tr\left((\T{W}'-\T{W})\star(\nabla_{\T{W}}{\cal H}(\T{W}))^\top\right) \\
&\stackrel{(a)}{=} \cH(\T{W}') - \cH(\T{W}) -\mathrm{Tr}\left((\T{W}'-\T{W})\star(\log \T{W})^\top\right) \\
&= \hcH(\widehat{\T{W}}') - \hcH(\widehat{\T{W}}) \nonumber \\
& \hspace{7mm} - \mathrm{Tr}\left(\blkd(\widehat{\T{W}}^\prime-\widehat{\T{W}})[\log(\blkd( \widehat{\T{W}}))]^\dagger\right) 
\end{align*}
\textbf{\emph{where (a) follows from the Lemma \ref{lem:grad} in the Appendix and the fact that $$\nabla_{\widehat{\T{W}}} \hat{\cH}(\widehat{\T{W}}) = \rshpT(\log(\blkd( \widehat{\T{W}})))\,\, ,$$ see  \cite{Lewis:1996ko},\cite{Degaldo09}}}.

\section{Online prediction for tensors: Problem set-up and main results}
\label{sec:prob_setup}
%

For the problem of online tensor prediction, the \emph{complexity structure} that we impose on the data tensor stems from the $(\beta,\tau)$ decomposability construction found in \cite{Hazan2012:un} to express ordinary matrices in a positive definite form. Let's begin by defining the original matrix decomposition therein.

\begin{definition}
Let $\M{A}\in \mathbb{R}^{m\times n}$ be any real matrix. The \emph{symmetrization} of $\M{A}$, $\mathrm{sym}(\M{A})$, is defined as 
\[
\mathrm{sym}(\M{A}) =
\begin{pmatrix}
0 & \M{A} \\ \M{A}^T & 0
\end{pmatrix}
\]
\end{definition}

\begin{definition}
Let $p$ be the dimension of $\mathrm{sym}(\M{A})$. Then $\M{A}$ is \textit{$(\beta,\tau)$-decomposable} for real numbers $\beta$ and $\tau$ if there exist positive-semidefinite matrices $\M{P},\M{N} \in \mcr{S}^{p\times p}_+$ such that

\begin{description}
\itemsep0em
\item[(1)] $\mathrm{sym}(\M{A}) = \M{P} - \M{N}$
\item[(2)] $\forall i, \M{P}(i,i),\M{N}(i,i) \leq \beta$
\item[(3)] $\mathrm{Tr}(\M{P}) + \mathrm{Tr}(\M{N}) \leq \tau$
\end{description}\vspace{-3mm}
\end{definition}

It turns out the notion of $(\beta,\tau)$-decomposability is tightly related to the max norm and nuclear norm of $\M{A}$, making it simple to find suitable decomposition parameters for any class of matrices. More precisely, the least possible $\tau$ used to decompose a matrix $\M{A}$ is equal to $2\|\M{A}\|_*$, and the least possible $\beta$ is $\frac{1}{2}||\M{A}||_\infty$ \cite{Hazan2012:un}, where $||\cdot||_*$ and $||\cdot||_\infty$ denote the matrix nuclear norm and $\ell_{\infty}$-norm. 

We will now extend this notion to tensors. 

\subsection{The class of \texorpdfstring{$(\V{\beta},\V{\tau})$}{(B,t)}-decomposable tensors}
\begin{definition}
Let $\T{A} \in \mathbb{R}^{m\times n \times d}$ be any tensor. We say that $\T{A}$ is \emph{$(\V{\beta},\V{\tau})$-decomposable} for $\V{\beta}$,$\V{\tau} \in \Real^d$, if $\forall k \in [d]$, $\widehat{\M{A}}^{(k)}$ is $(\beta(k),\tau(k))$ decomposable. Additionally, we say a set $\mcr{S} \in \mathbb{R}^{m \times n \times d}$ is $(\V{\beta},\V{\tau})$-decomposable if each tensor in $\mcr{S}$ is $(\V{\beta},\V{\tau})$-decomposable.
\end{definition}

Note the distinction between the tensor and matrix case: decomposability of a tensor is determined in the Fourier domain. Indeed, $(\V{\beta},\V{\tau})$-decomposability implies disjoint, face-wise complexity restrictions on the Fourier tensor, which in turn captures the number of non zero singular tubes in the t-SVD of the tensor.


We now state the central problem addressed by this paper in the box below\footnote{We enforce $||\V{\beta}||_1 \geq 1$ for analytical convenience, and as noted in \cite{Hazan2012:un} this is only a mild restriction.}. The goal of Online Tensor Prediction is to minimize regret, which is defined as
\[
\mathrm{Regret} \triangleq
 \sum_{t=1}^T l_t(\T{A}_t(i_t,j_t,k_t)) - \arg\min_{\T{U} \in \mcr{S}}\sum_{t=1}^T l_t(\T{U}(i_t,j_t,k_t))
\] 


\begin{figure}
    \begin{minipage}{0.5\textwidth}
 
\begin{framed}
\begin{center}

{\fontfamily{cmtt}\selectfont Online Tensor Prediction}
\end{center}
\textbf{parameters:} $\V{\beta} \succeq 0$ with $||\V{\beta}||_1\geq 1$, $\V{\tau} \succeq 0$, $G \geq 0$, $m$, $n$, $d$ \\
\textbf{input:} A $(\V{\beta},\V{\tau})$-decomposable set $\mcr{S} \subseteq [-1,1]^{m\times n \times d}$ \\
\textbf{for} $t = 1,2,3,...T$ \\
\hspace{1cm} adversary supplies indices $(i_t,j_t,k_t) \in [m]\times[n]\times[d]$ \\
\hspace{1cm} learner predicts $p_t = \T{A}_t(i_t,j_t,k_t)$ from a maintained tensor $\T{A}_t \in \mcr{S}$ \\
\hspace{1cm} adversary supplies a convex, $G$-Lipschitz function $l_t: [-1,1] \rightarrow \mathbb{R}$ \\
\hspace{1cm} learner suffers loss $l_t(p_t)$ \\
\textbf{end for}
\end{framed}
\end{minipage}
\end{figure} 
Given the set up, our main result is summarized by the following theorem. 
\begin{theorem} \textbf{[Main Result]}
\label{thm:TEG_regret}
There exists an algorithm for Online Tensor Prediction with regret bounded by
\[
Regret \leq 2G\sqrt{\log(2p)T(\sum_{k=1}^{d}\tau(k))(\sum_{k=1}^{d}\beta(k))}
\]
where $p$ is the dimension of each $\mathrm{sym}(\widehat{\M{A}}^{(k)})$.
\end{theorem}

\emph{\textbf{Proof outline}}: The algorithm is given in Section \ref{sec:OTEG}. For this algorithm We find the regret bound for our algorithm by linearly approximating the loss functions and applying a linear regret bound to obtain Theorem~\ref{thm:TEG_regret}. To find the linear bound, we rely on the $(\V{\beta},\V{\tau})$-decomposability of the learning set to transfer the problem to a positive-definite domain, allowing us to use the Tensor Exponentiated Gradient algorithm, which is derived below. The complete proof can be found in the Appendix. \emph{Several important ingredients in the proof rely on some key results on gradient calculus in complex Hilbert spaces.}
\subsection{Tensor Exponentiated Gradient (TEG) Descent algorithm}
\label{sec:TEG}
We first derive an online algorithm for learning of symmetric PD tensors.

\textbf{TEG set-up}: On each round $t$, we are given an \emph{instance tensor} $\T{X}_t \in \Real^{N \times N \times d}$, the learner predicts the tensor $\T{W}_t$ from a convex set $\mcr{W} \subseteq \mcr{S}^{N \times N \times d}_{++}$, receives a convex loss function $L_t :\mcr{W} \rightarrow \mathbb{R}$, and suffers the loss $L_t(\T{W}_t)$. 

For all $t$, we assume that the gradient  $\nabla_{\T{W}}L_t$, which is a tensor, is well defined and face-wise symmetric. In addition we assume there exists a function $\widehat{L}_t(\widehat{\T{W}})$ which is convex in $\widehat{\T{W}}$ with $L_t(\T{W}) = \widehat{L}_t(\widehat{\T{W}})$. Clearly, $\widehat{L}_t(\widehat{\T{W}}) = L_t(\ifft(\widehat{\T{W}}, [\,], 3))$. 

Following \cite{Tsuda2005wc} we now derive the Tensor Exponentiated Gradient update, which is equivalent to a standard Matrix Exponentiated Gradient with block-diagonal matrices --
\begin{align*}
\T{W}_{t+1} 
&= \arg\min_{\T{W} \in \mcr{W}} \Delta_H(\T{W},\T{W}_t) + \eta \langle \T{W}, \nabla_{\T{W}}L_t(\T{W}_t) \rangle
\end{align*}
where $\eta$ is the \emph{learning rate}. Equivalently in the Fourier domain we have, 
\begin{align}
\label{eq:TEG_opt1}
\widehat{\T{W}}_{t+1}= & \arg\min_{\widehat{\T{W}} \in \widehat{\mcr{W}}} \Delta_{\widehat{H}}(\widehat{\T{W}},\widehat{\T{W}}_t) \nonumber \\ 
 & \hspace{1mm} + \eta\mathrm{Tr}\left(\blkd(\widehat{\T{W}}) \,[\blkd (\nabla_{\widehat{\T{W}}} \widehat{L}_t(\widehat{\T{W}}_t))]^\dagger\right),
\end{align}
where $\widehat{\mcr{W}}$ denotes the set of tensors obtained by taking the Fourier transform of each tensor in $\mcr{W}$. From \cite{Tsuda2005wc} and \cite{Hazan2012:un}, we know that the closed form solution to optimization problem in Equation~(\ref{eq:TEG_opt1}) is given by a projected exponentiated gradient descent of Equation (\ref{eq:TEG_opt2}).
\begin{figure*}
\begin{align}
\label{eq:TEG_opt2}
\widehat{\T{W}}_{t+1} = \arg \min_{\widehat{\T{W}} \in \widehat{\mcr{W}}} \Delta_{\widehat{H}}\left(\widehat{\T{W}},
	\rshpT\left(
		\exp\left(
			\log( \blkd(\widehat{\T{W}}_t)) - 
			\blkd\left(\eta			\nabla_{\widehat{\T{W}}} \widehat{L}_t(\widehat{\T{W}}_t)\right)
		\right)
	\right)
\right)
\end{align}
\vspace{-5mm}
\end{figure*}

Recalling that $\T{W}_t = \ifft(\widehat{\T{W}}_t, [\,], 3)$, we obtain the Tensor Exponentiated Gradient algorithm for online learning of symmetric PD tensors. Note that the optimization in the equation above can be parallelized, since $\exp$ and $\log$ of a block-diagonal matrix are computed block-by-block.

\subsection{An Algorithm for Online Tensor Prediction}
\label{sec:OTEG}
We begin with a simple construction that lets us represent a $(\V{\beta},\V{\tau})$-decomposable tensor as a positive-definite tensor. Let $\mcr{S} \subseteq [-1,1]^{m\times n \times d}$ be a $(\V{\beta},\V{\tau})$-decomposable set. For $\T{A} \in \mcr{S}$, let $\widehat{\T{P}}$,$\widehat{\T{N}} \in \mcr{S}^{p \times p \times d}_{+}$ be the Fourier tensors such that $\mathrm{sym}(\widehat{\M{A}}^{(k)}) = \widehat{\M{P}}^{(k)} - \widehat{\M{N}}^{(k)}$. We define $\widehat{\phi} : \mcr{S} \to \mathbb{C}^{2p \times 2p \times d}$ face-wise as
\[
\widehat{\phi}(\T{A})^{(k)} = \begin{pmatrix}
\widehat{\M{P}}^{(k)} & 0 \\ 0 & \widehat{\M{N}}^{(k)} \end{pmatrix}
\]
The set of all such $\widehat{\phi}$ constructions over the learning set  $\mcr{S}$ will be contained in the convex set $\widehat{\mcr{W}}$, defined by
Equation~(\ref{eq:Kset}). 

\begin{figure*}
\begin{align}
\widehat{\mcr{W}} = &\left\{ \ \widehat{\T{W}} \in \mathbb{C}^{2p \times 2p \times d} : \forall k, \widehat{\M{W}}^{(k)} \succeq 0, \forall k \forall i, \widehat{\M{W}}^{(k)}(i,i) \leq \beta(k) \right. \notag\\
& \ \ \left.\forall k, \mathrm{Tr}(\widehat{\M{W}}^{(k)}) \leq \tau(k),\,\, \forall (i,j,k) \in [m]\times[n]\times[d]: P_{\widehat{\T{W}}}(i,j,k) \in [-1,1]  \right\} \label{eq:Kset}
\end{align}
\vspace{-5mm}
\end{figure*}

\begin{figure}
    \begin{minipage}{0.5\textwidth}
\begin{algorithm}[H]
\caption{Tensor Exponentiated Gradient for Online Tensor Prediction (OTEG)}
\label{alg:OTEG}
\begin{algorithmic}
\STATE
\textbf{input}: $m$,$n$,$d$,$G$,$\V{\beta}$, $\V{\tau}$ \\
\textbf{set}: $p = m+n$, $N = 2p$, $\widehat{\mcr{W}}$ as in (\ref{eq:Kset}), $\forall k: \gamma(k)=4G^2$ \\
$\eta = \sqrt{\frac{\log N\sum_{k=1}^d\tau(k)}{ T\sum_{k=1}^d \gamma(k)\beta(k)}}$ \\
\textbf{initialize}: $\forall k: \widehat{\M{W}}_1^{(k)} = \frac{\tau(k)}{N}\M{I}$
\FOR{$t=1,2,3,...$}
\STATE
Receive triplet of indices $(i_t,j_t,k_t) \in [m]\times[n]\times[d]$ \\
Predict $p_t = P_{\widehat{\T{W}}}(i_t,j_t,k_t)$ \\
Receive $G$-Lipschitz, convex loss function \\ \ \ \ \ \ $l_t : [-1,1] \rightarrow \mathbb{R}$ and suffer loss $l_t(p_t)$ \\
Calculate the subderivative $g$ of $l_t$ at $p_t$ \\
Construct loss tensor $\widehat{\T{L}}_t
= \nabla_{\widehat{\T{W}}}\widehat{L}_t(\widehat{\T{W}}_t)$\\
Update $\widehat{\T{W}}_{t+1}$ by solving (\ref{eq:TEG_opt1})
\ENDFOR
\end{algorithmic}
\end{algorithm}
\end{minipage}
\end{figure}

Where $P_{\widehat{\T{W}}}(i,j,k) = [\ifft(\widehat{\T{P}}-\widehat{\T{N}})](i,j+m,k) \triangleq p_t$ is the so-called ``prediction" operator, which extracts $\T{A}_t(i,j,k)$ from the its positive-definite embedding $\widehat{\phi}(\T{A}_t)=\T{W}_t$. In our case, $\T{X}_t$ is a tensor that encodes the indices $(i_t,j_t,k_t)$, and we restrict our loss functions to the form $L_t(\T{W}_t) = l_t(p_t)$. Note that since $\widehat{\mcr{W}}$ is composed of Fourier-domain tensors, the tensor entries will be complex in general, but with real face-wise diagonal entries and trace since the frontal faces are Hermitian.

In order to apply Tensor Exponentiated Gradient, we must compute $\nabla_{\widehat{\T{W}}}l_t(P_{\widehat{\T{W}}_t}(i_t,j_t,k_t))$, the gradient of the current loss with respect to $\widehat{\T{W}}$. We have 
\begin{align*}
&\nabla_{\widehat{\T{W}}}l_t(P_{\widehat{\T{W}}_t}(i_t,j_t,k_t)) \notag \\
& \stackrel{(a)}{=} \nabla_{p}l_t(p_t)\nabla_{\widehat{\T{W}}}P_{\widehat{\T{W}}_t}(i_t,j_t,k_t) \notag \\
& \stackrel{(b)}{=} g\nabla_{\widehat{\T{W}}}P_{\widehat{\T{W}}_t}(i_t,j_t,k_t) 
\end{align*}
where $(a)$ follows from the chain rule, and $(b)$ defines $g = \nabla_{p}l_t(p_t)$.

We see the gradient is split into two components: the ``time domain gradient" ($g$), and the ``Fourier domain gradient" ($\nabla_{\widehat{\T{W}}}P_{\widehat{\T{W}}_t}(i_t,j_t,k_t)$), which will be a complex gradient of a real-valued function (see \cite{Degaldo09}).

\begin{figure}
    \begin{minipage}{0.5\textwidth}
\begin{align*}
\widehat{\T{L}}_t(i,j,:)= 
\begin{cases}
g\M{F}(:,k) & \mathrm{if} \ (i,j) = (i_t,j_t+m) \\
-g\M{F}(:,k) & \mathrm{if} \ (i,j) = (i_t+p,j_t+m+p) \\
g\overline{\M{F}}(:,k) & \mathrm{if} \ (i,j) = (j_t+m,i_t) \\
-g\overline{\M{F}}(:,k) & \mathrm{if} \ (i,j) = (j_t+m+p,i_t+p)\\
0 & \mathrm{otherwise}
\end{cases}
\end{align*}
\end{minipage}
\vspace{-5mm}
\end{figure}

By straightforward calculation we find an explicit expression of the gradient (LHS), which is built by arranging copies of one column of the DFT matrix $\M{F}$ and its complex conjugate along tubes in the third dimension. Note that each $\widehat{\M{L}}_t^{(k)}$ is Hermitian, as required by the Tensor Exponentiated Gradient update, and that $(\widehat{\M{L}}_t^{(k)})^2$ is a matrix with four copies of $g^2$. In fact, since $g \leq G$, we can now write a $\V{\gamma}$ constraint for $\widehat{\T{L}}_t$ -- $\forall k, \gamma(k) = 4G^2$.


Based on this development our algorithm for Online Tensor Prediction is given in Algorithm~\ref{alg:OTEG}. For convenience, we assume that the tensor with faces $\widehat{\M{W}}^{(k)} = \frac{\tau(k)}{N}\M{I}$ is in $\widehat{\mcr{W}}$.

\section{Experimental validation}
\label{sec:sim}
\subsection{Generation of \emph{semi-synthetic} temporal rating dataset}
For experimentation we generated \textbf{coupled-time dynamics} in recommendation systems with social network interaction. Specifically, we generated a \emph{semi-synthetic} data set of user ratings evolving over time in the following manner. True rating data set was taken from the Movie Lens \cite{movie_lens} movie rating observation data and then truncated to only 150 users and 100 movies in order to manage the size of the simulation. In this data set, every user had rated at least 20 movies, but not all 100, so we used a low rank completion method \cite{Recht:2010:GMS:1958515.1958520} to complete the initial rating matrix. We then mapped the users to a social network taken from the Stanford Network Analysis Project (SNAP) \cite{snap}. Our network was a subset of an undirected graph with 1034 nodes and 88234 edges and an average clustering coefficient of 0.6055. We simply used the first 150 nodes in the network to represent our users. Then we took the initial complete rating matrix and evolved the ratings of the users by using the following two influence models as dictated by the social network.

\begin{itemize}

\item \textbf{Dataset A: Evolution with neighborhood influence}- For this data set we used the following evolution model for the ratings tensor $\T{M}$. At each time epoch $s$ the users update their ratings according to, 

\begin{align}
\T{M}(u,m,s)= a(s)\, \mbox{Neighbor} \,\, +\,\, b(s)\, \mbox{Self}
\end{align}

where,
\begin{align}
 \mbox{Neighbor} =\frac{\sum_{u' \in {\cal N}(u)} \T{M}(u',m,s-1)}{|{\cal N}(u)|}\\
\mbox{Self}= \frac{\T{M}(u,m,s-1)+ {\tt rand}([1:5])}{2}
\end{align}

and where $ {\cal N}(u)$ denotes the set of neighbors (in the undirected graph) of user $u$ and the function ${\tt rand}([1:5])$  outputs a random rating between 1 and 5. In this evolution $a(s) \in [0,1]$ are random constants and $b(s) =1-a(s)$. The reason for selecting this evolution pattern is based on the following. We know that the new rating should be a combination of neighborhood influence, past opinions (personal rating at previous time), and a random influence or self-innovation. The amount of influence that a user's friends had on his rating should be variable between different users, so the weighting of the neighborhood influence was randomized. Accordingly, user's personal influence was then weighted accordingly to have a total weighting of 1. These kinds of models have been recently studied in \cite{Jadbabaie}. We call this data as Dataset A. The size of this dataset is $150\times 100 \times 20$. 

\item \textbf{Dataset B: Evolution with neighborhood influence with stubborn rating dynamics}- For this data set, one additional property was added. In order to parallel what we believe is the norm in the real world, once a user has rated a particular movie a 5, i.e. the top rating, his/her rating for that movie cannot decrease. This property holds for a user that obtains a rating of five at any point in the simulation, not only the initial time. The rest of the dynamics is same as in the previous case. We call this data as Dataset B. The size of this dataset is $150\times 100 \times 25$. 
\end{itemize}

\textbf{Note}: that the time steps for the online algorithm (i.e. the sequential plays) and the time steps in evolution of the ratings patterns are conceptually different and should not be confused with each other. In particular at any step in the algorithm one can play any value in the data cube. One can also consider another scenario where there are many sequential plays for each rating evolution step with the indices in the sequential play restricted to the evolution data cube dimensions so far. This will not affect the algorithm (since the index drawing is adversarial in nature). Further note that in the evaluation below we will not use the knowledge of the social network and the evolution models. The network and evolution model is just to generate datasets for testing the proposed methods.

\subsection{Evaluation of the proposed algorithm}

We simulate online learning of these datasets as follows: at time $t$, a rating for a particular user-movie-time is sampled at random from a uniform distribution among the set of indices which have not yet been played, and this index is played as $y_t$. We apply two algorithms in this setup: (1) a partial implementation of OTEG, and (2) a standard follow-the-regularized-leader approach with  tensor-nuclear-norm regularization. For both experiments, $T= 20$\% of the data cube, and $l_t(p_t)=(y_t-p_t)^2$.

\begin{figure*}
\centering \makebox[0in]{
    \begin{tabular}{cc}
    \vspace{-30mm}\\
    \includegraphics[height = 4in, width = 3in]{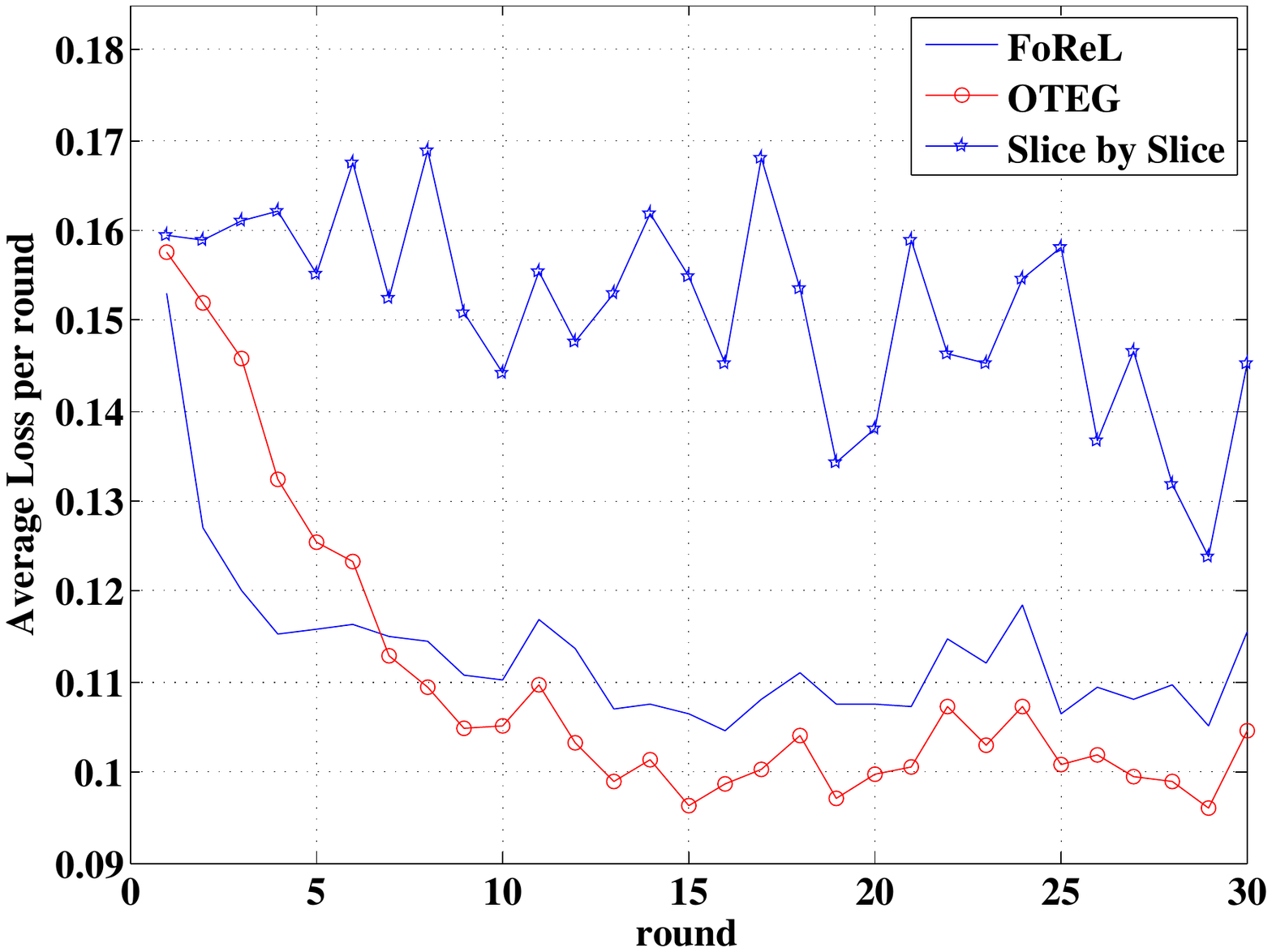} &
      \includegraphics[height = 3.8in, width = 3in]{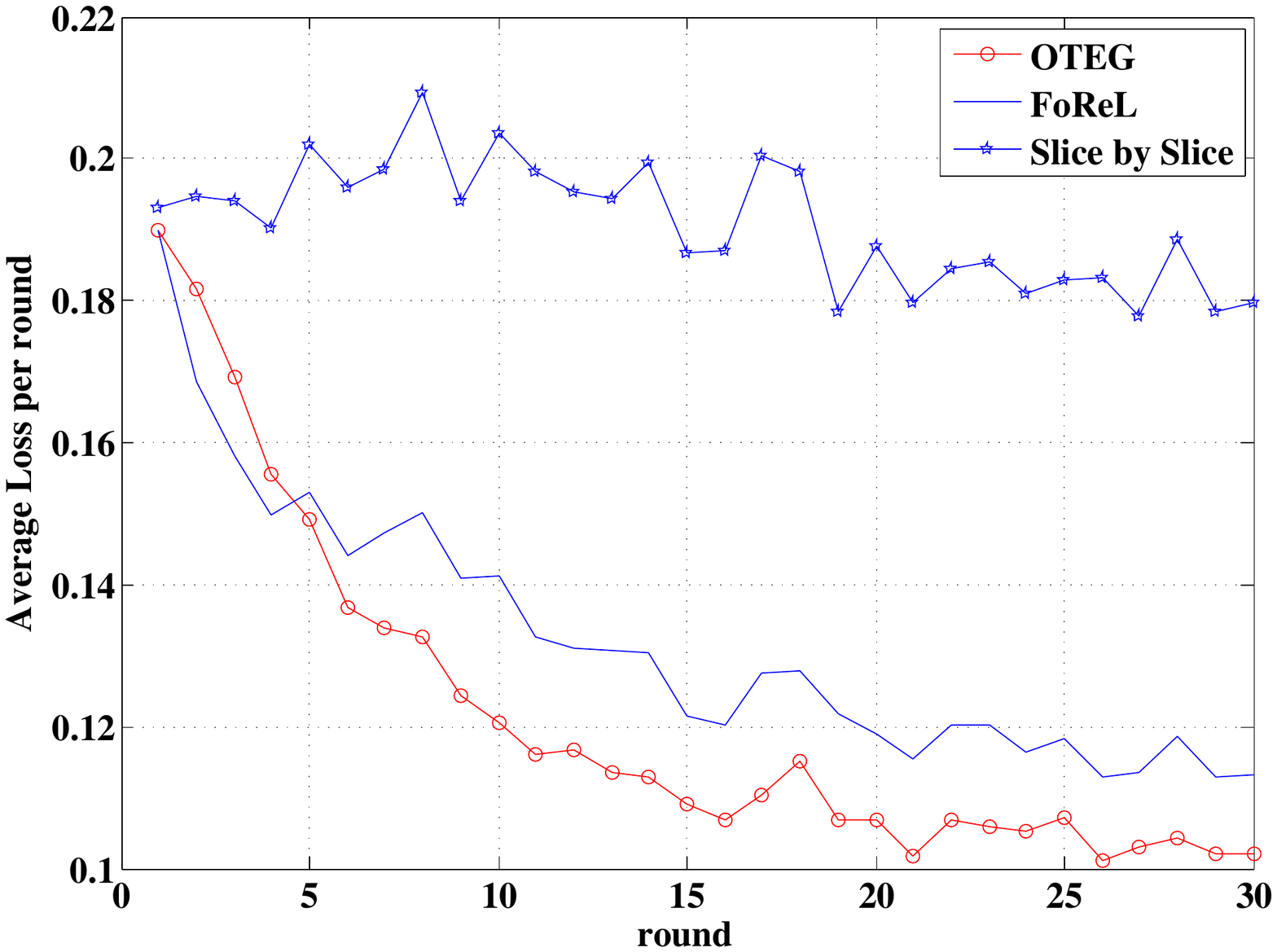}\\
    \end{tabular}}
    \vspace{-25mm}
\caption{\textbf{Left}: Loss plots for OTEG, FoReL and naive methods after 60000 iterations for Dataset A. \textbf{Right}: Loss plots for OTEG, FoReL and naive methods after 75000 iterations for Dataset B.}
\label{fig:loss1}
\end{figure*} 

\textbf{OTEG Experiment} - Based on the collaborative filtering example in \cite{Hazan2012:un}, we choose $\beta(k)=\sqrt{n+m}$ and $\tau(k) \approx 2||\M{\widehat{M}}^{(k)}||_*$. A small uniform random noise in $[0,5]$ is added to $\tau(k)$ to simulate imperfect a-priori knowledge of the tensor nuclear norm of $\T{M}$. Since a full implementation of OTEG requires solving a semi-definite program at each time step--which is computationally expensive and tedious to program--we opt to simplify the projection step of the algorithm. Specifically, instead of projecting onto $\widehat{\mcr{W}}$, we project onto $\{\widehat{\T{W}}|\mathrm{Tr}(\blkd{(\widehat{\T{W}})})\leq\tau\}$ via trace normalization (\emph{a la} Algo. 1 in \cite{Tsuda2005wc}). This results in slower convergence, so to compensate the learning rate dictated by Algorithm \ref{alg:OTEG} is increased by a factor of 8, i.e. $\eta = 4\sqrt{\frac{\log (2(m+n))\sum_{k=1}^d\tau(k)}{ TdG^2\sqrt{n+m}}}$. The lack of full projections also implies that $p_t$ is not necessarily in $[-1,1]$, so we cannot analytically calculate a Lipschitz constant for $l_t$. Instead, $G$ is the current maximum value of $|l_t'(p_t)|=|2(y_t-p_t)|$ and is continuously updated (along with $\eta$) as the experiment progresses. 

\textbf{FoReL Experiment}--FoReL  is a standard approach to online convex optimization \cite{ShalevShwartz:2012dz}. On each round $t$, we perform the update 
\[
\vspace{-1mm}
\T{W}_{t+1} = \arg \min_{\T{W}} ||\mcr{P}_t(\T{W} - \T{M})||_{F}^{2} + \eta \sum_{k=1}^{d} ||\widehat{\M{W}}^{(k)}||_{*}
\]
Where $\mcr{P}_t$ is a sampling operator that zeros out entries of a tensor corresponding to indices not yet sampled. This algorithm is implemented using FISTA \cite{Beck:2009gh} using 5 gradient descent iterations per update. We use the learning rate $\eta = \frac{B}{G\sqrt{T}}$, where $B=1.1||\T{M}||_F$ is an upper bound on $||\T{M}||_F$. $G$ is calculated in the same fashion as the OTEG experiment.

\textbf{Completion slice by slice} - We further compared our algorithm against the naive approach of slice by slice tensor completion where each slice was completed independently of each other (in the original domain).

The results of the three experiments are shown in Figure~\ref{fig:loss1} for Dataset A and Dataset B respectively. The loss plots show a moving average of the value of $l_t(p_t)$, where the $T$ time intervals are divided into $R=30$ \emph{rounds} and we plot the average loss over each round. Note that while FoReL is better than OTEG in the first few rounds, OTEG seems to be slightly better than FoReL in the subsequent rounds. The reason for this is due to the fact that while solving for FoReL in each step we restrict the number of iterations to only 5 (for sake of reducing computation time). In other words we only partially solve the FoReL at each step. Also note that the slice by slice completion strategy is sub-optimal.


\textbf{OMEG Experiment}- A popular technique for tensor prediction is based on first flattening the tensor into a matrix followed by exploiting strategies for matrix completion. We will show here that such strategies are not necessarily optimal. We implement a version of the OMEG algorithm, \cite{Hazan2012:un}, using OTEG but with the third dimension set to 1. The data for comparison is a reduced size data generated in the same manner as Dataset B with dimensions $50 \times 60 \times 20$. For OMEG, the 3-D data cube was flattened to a 2-D matrix in the 3 possible ways or \emph{modes}, \cite{TomiokaSHK11}. The total number of plays for this set-up was chosen to be $T = 12,000$ which is about $20$\% of the data. Note that the error performance of the best possible mode flattening for OMEG is well below the OTEG performance. 
\begin{figure}
\centering \makebox[0in]{
    \begin{tabular}{c}
       \vspace{-25mm}\\
    \includegraphics[width=0.4\textwidth]{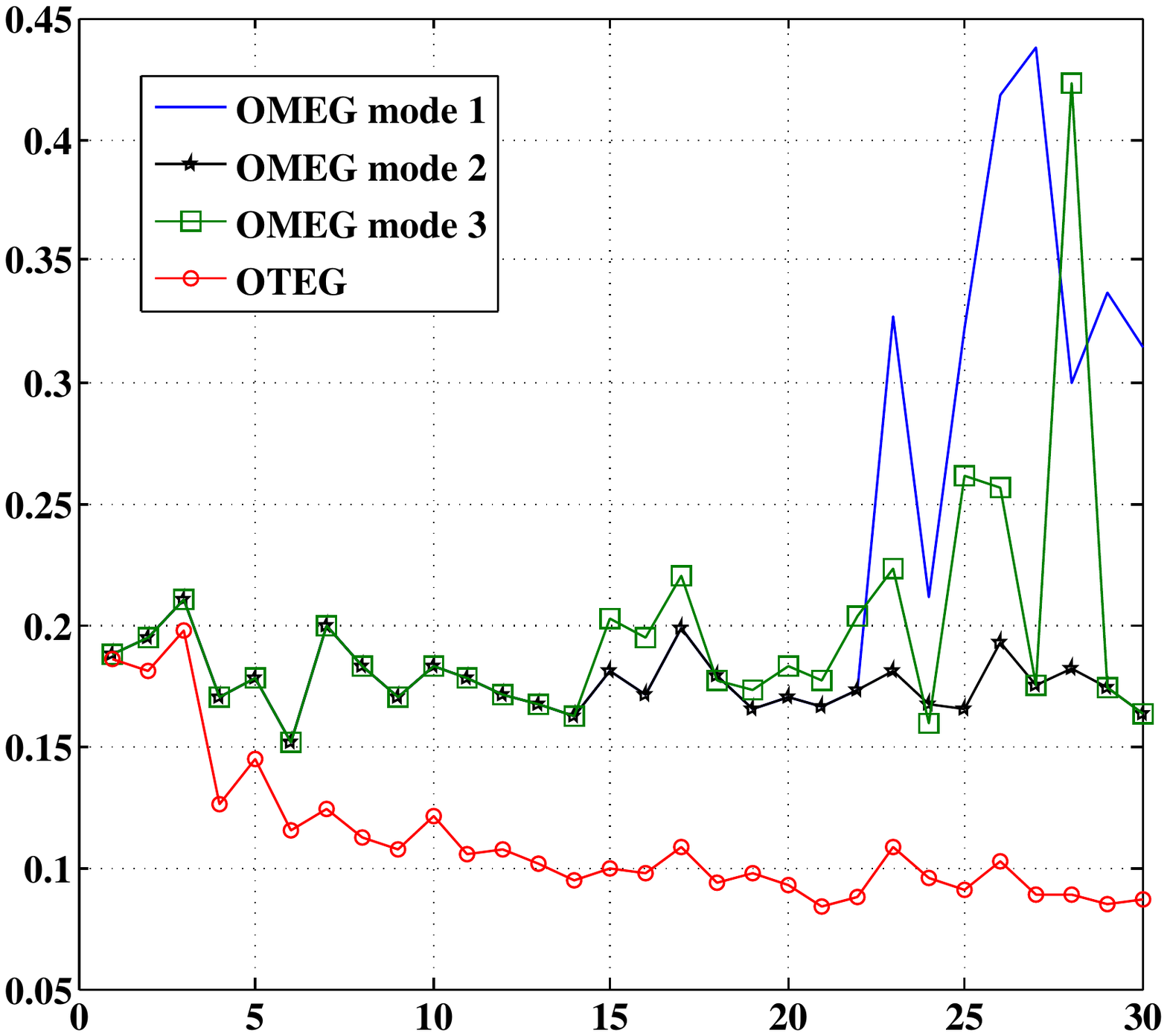}
    \end{tabular}}
    \vspace{-22mm}
\caption{Loss plots for OTEG  vs OMEG on a reduced size data cube. Note the superior performance of OTEG compared to OMEG}
    \vspace{-5mm}
\label{fig:loss_OMEG}
\end{figure}

\textbf{Discussion}-- Note the ``total memory'' approach of FoReL: every past sample is stored and used in calculating future updates. In contrast in OTEG implementation past plays are not stored and updates are calculated from only gradient information. While our implementation of OTEG stores all prior gradients, the full implementation theoretically requires only the latest gradient, which can be encoded with $g$ and $(i_t,j_t,k_t)$. Taking into account the storage of $\T{W}_t$, this implies worst case OTEG memory consumption\footnote{Some constant-factor memory gains can be achieved for OTEG by encoding $\widehat{\T{W}}_t$ with just half of the top-left and bottom-right blocks, due to structure of $\hat{\phi}(\T{A})$.} is $O((m+n)^2d)$, in contrast to $O(mnd+T)$ + Cost of storing past loss functions $l_t(\cdot)$ for FoReL. Note that in the current implementation of FoReL using FISTA one needs to recall all the past gradients (first order information only) of $\ell_t(\cdot)$ at each step. However, \textbf{since the the loss function is fixed for all time steps} for the current set of experiments, the memory cost of FoReL for our experiments is $O(mnd+T)$.   

%

\subsection{Simulation Results on video data}
\label{sec:sim}
\begin{figure}[h]
\centering \makebox[0in]{
    \begin{tabular}{ccc}
      \vspace{-14mm}\\
    \includegraphics[width=0.18\textwidth]{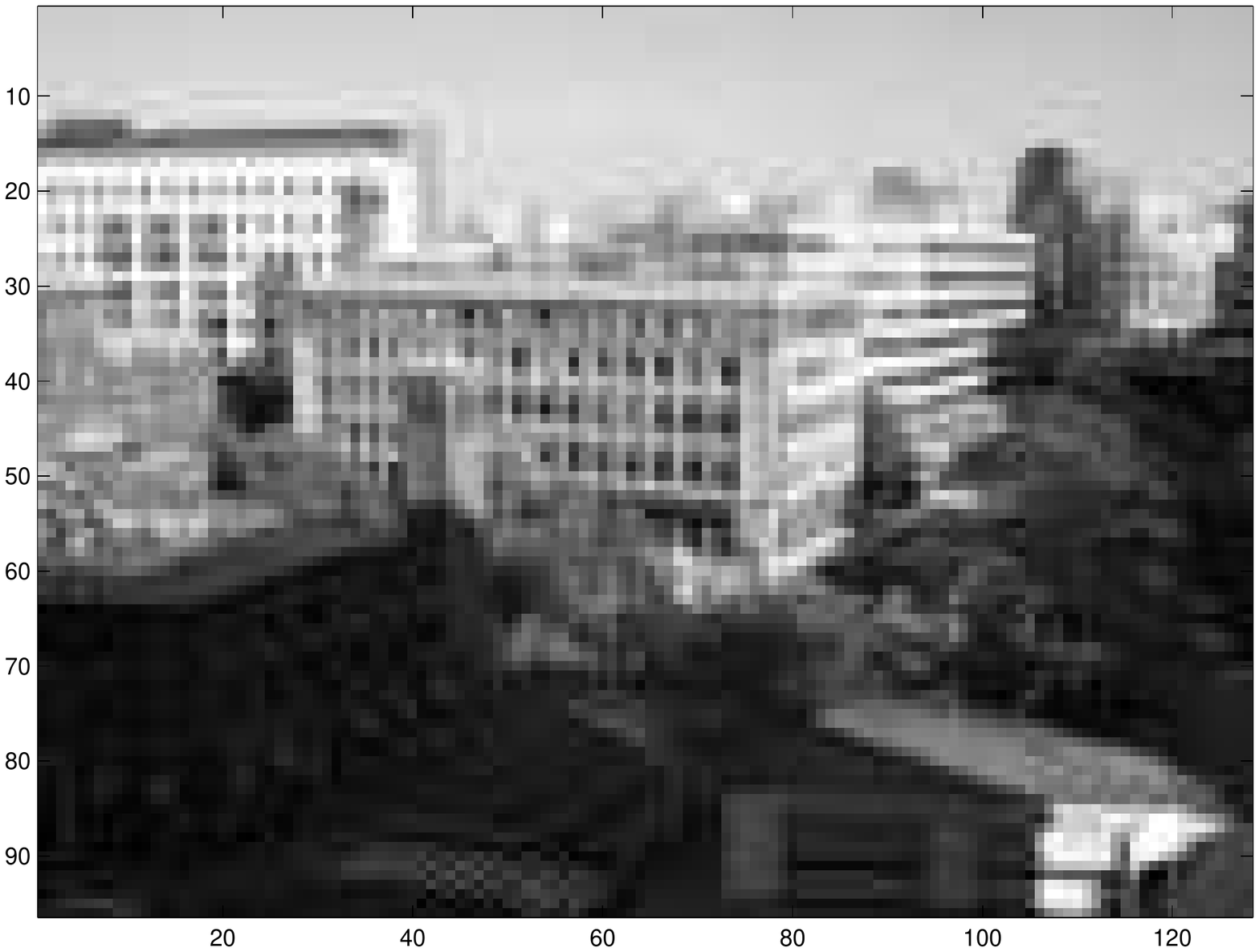} & \hspace{-5mm} \includegraphics[width=0.18\textwidth]{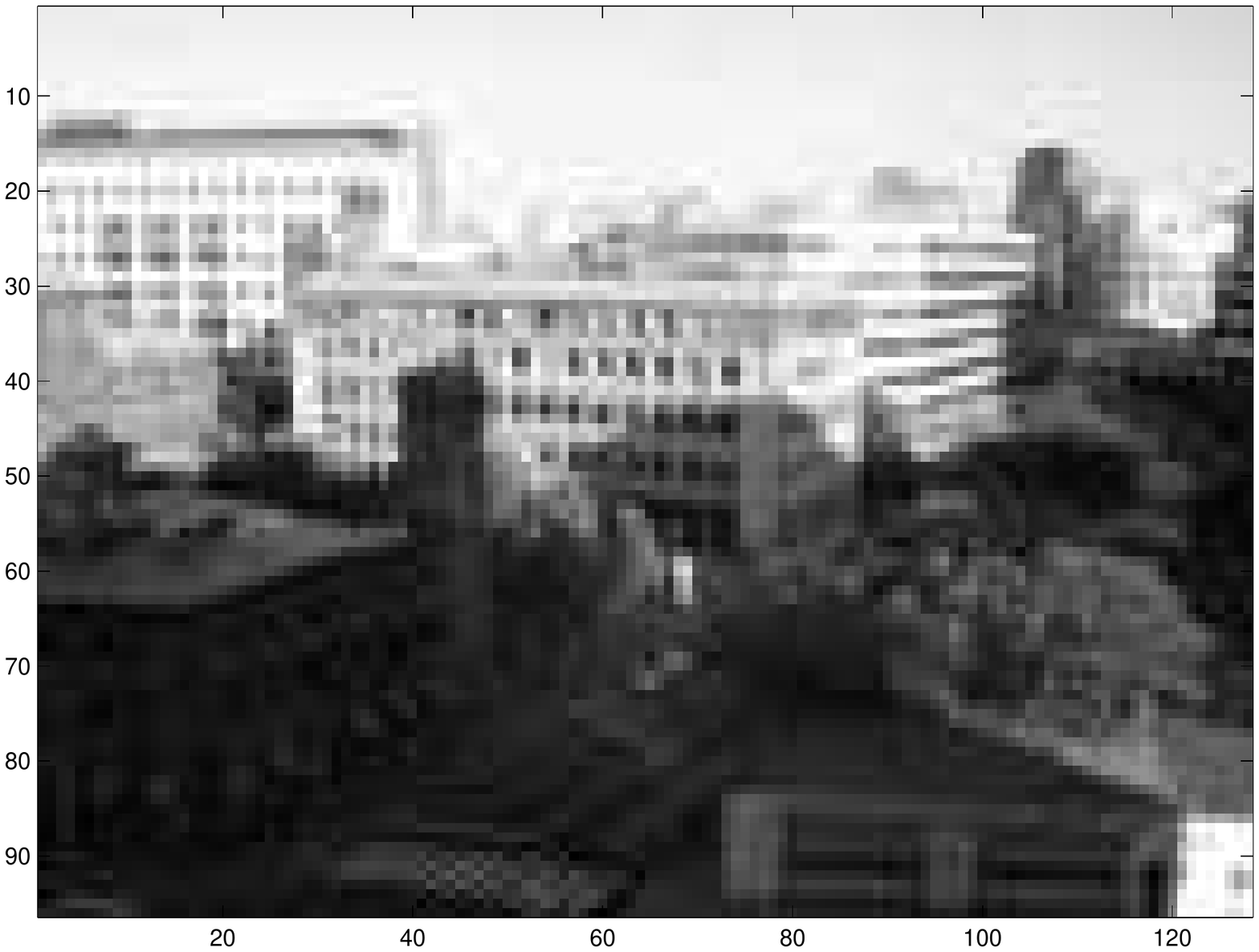} & \hspace{-5mm} \includegraphics[width=0.18\textwidth]{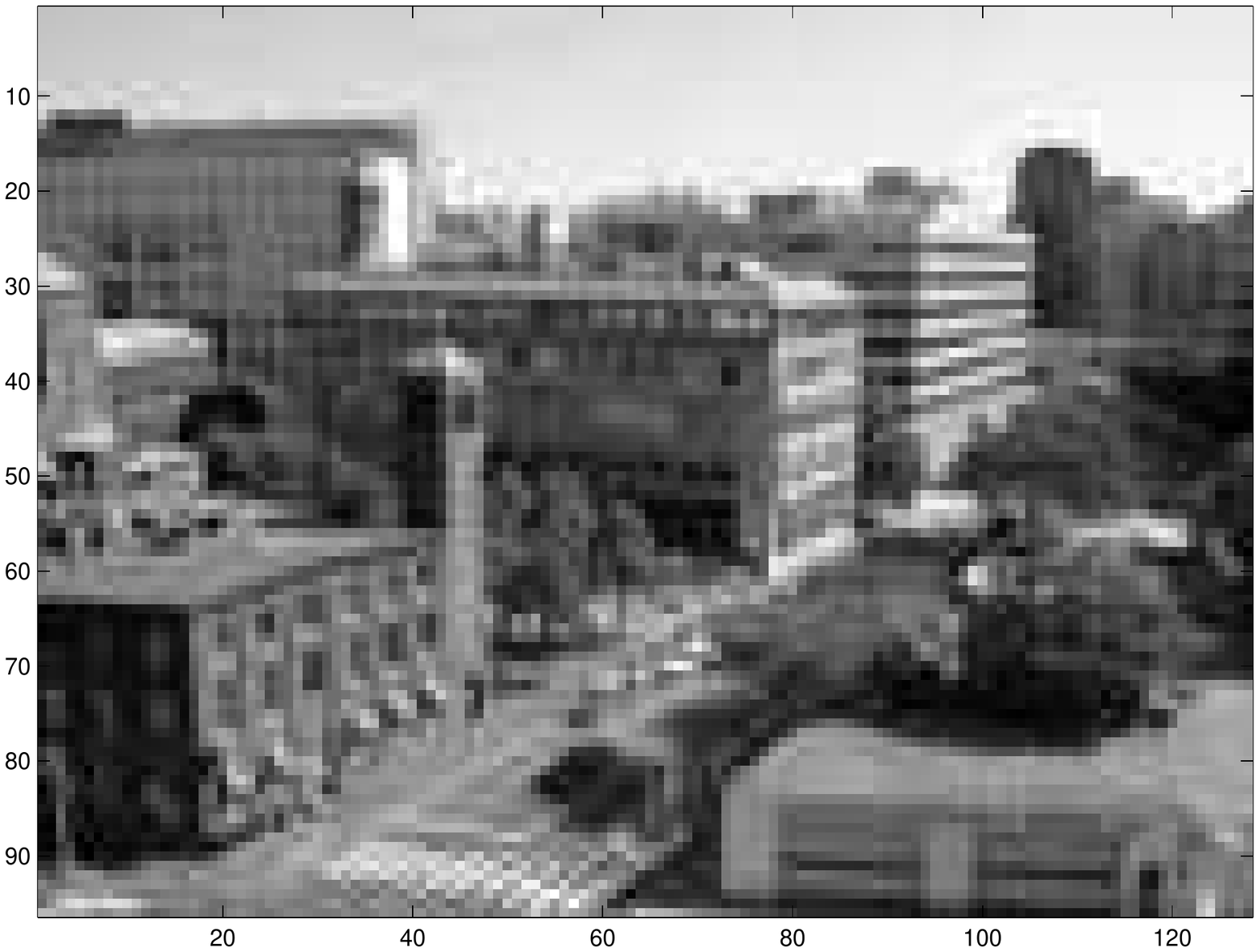}\\
    \vspace{-20mm}\\
       \includegraphics[width=0.18\textwidth]{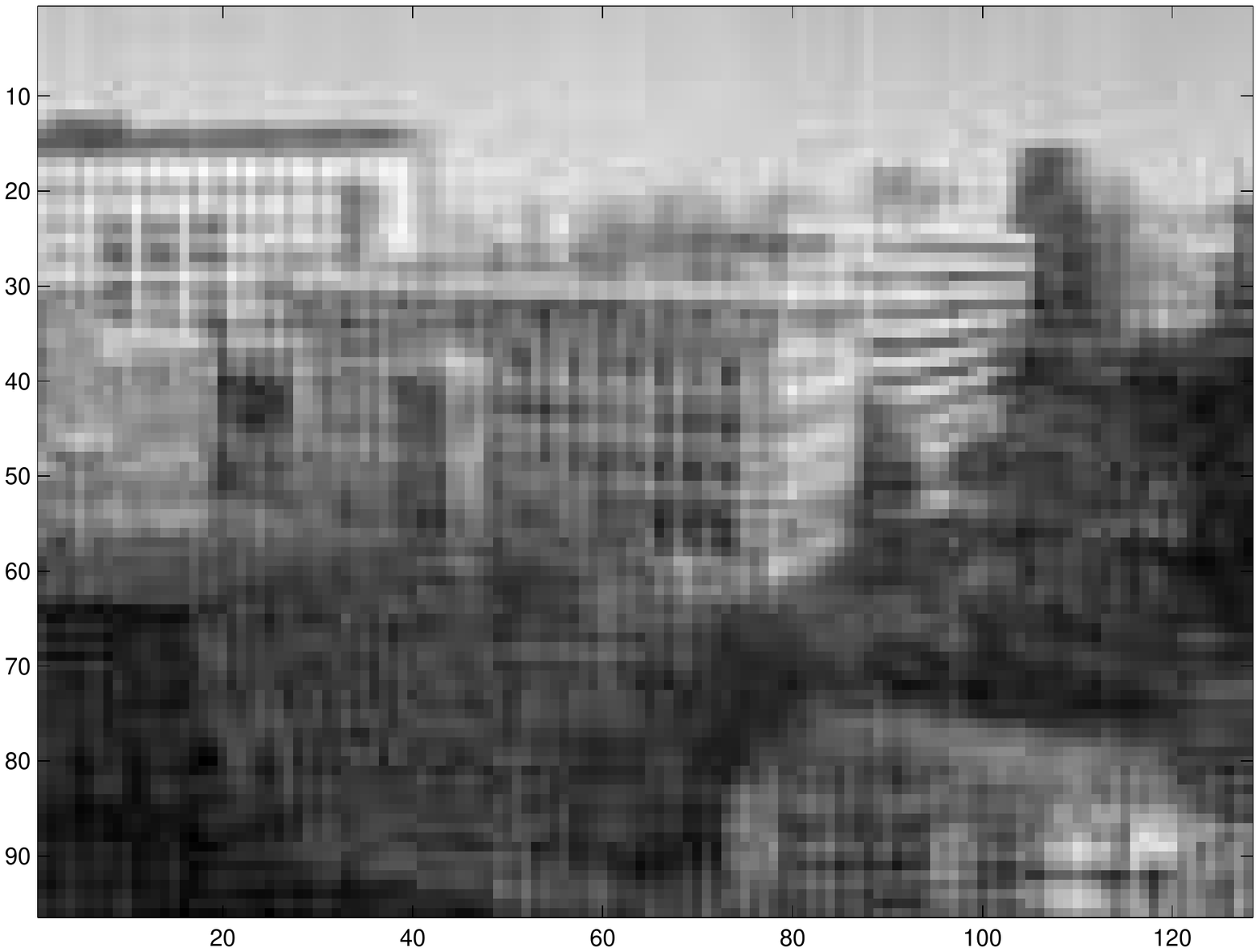} & \hspace{-5mm}  \includegraphics[width=0.18\textwidth]{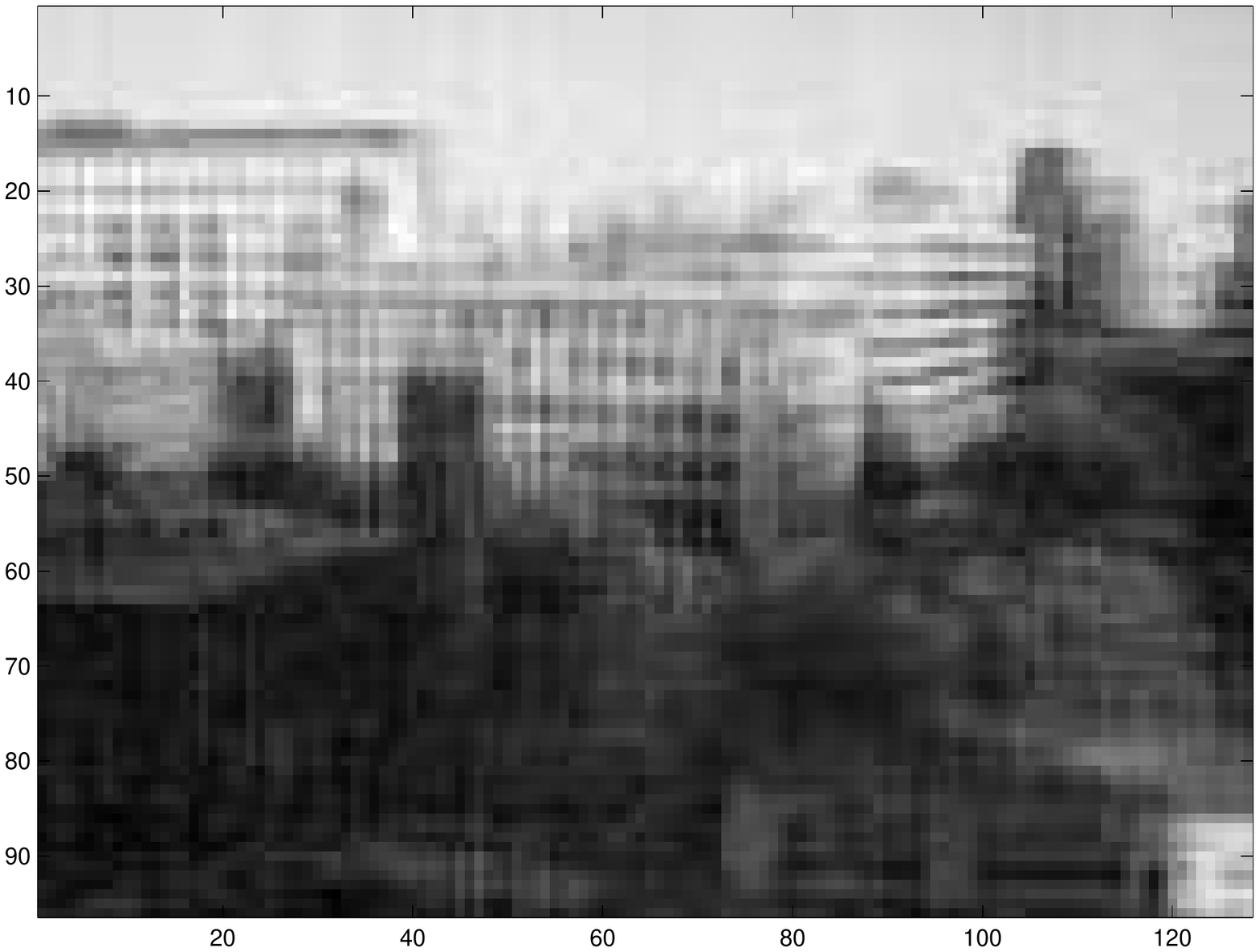}&  \hspace{-5mm} \includegraphics[width=0.18\textwidth]{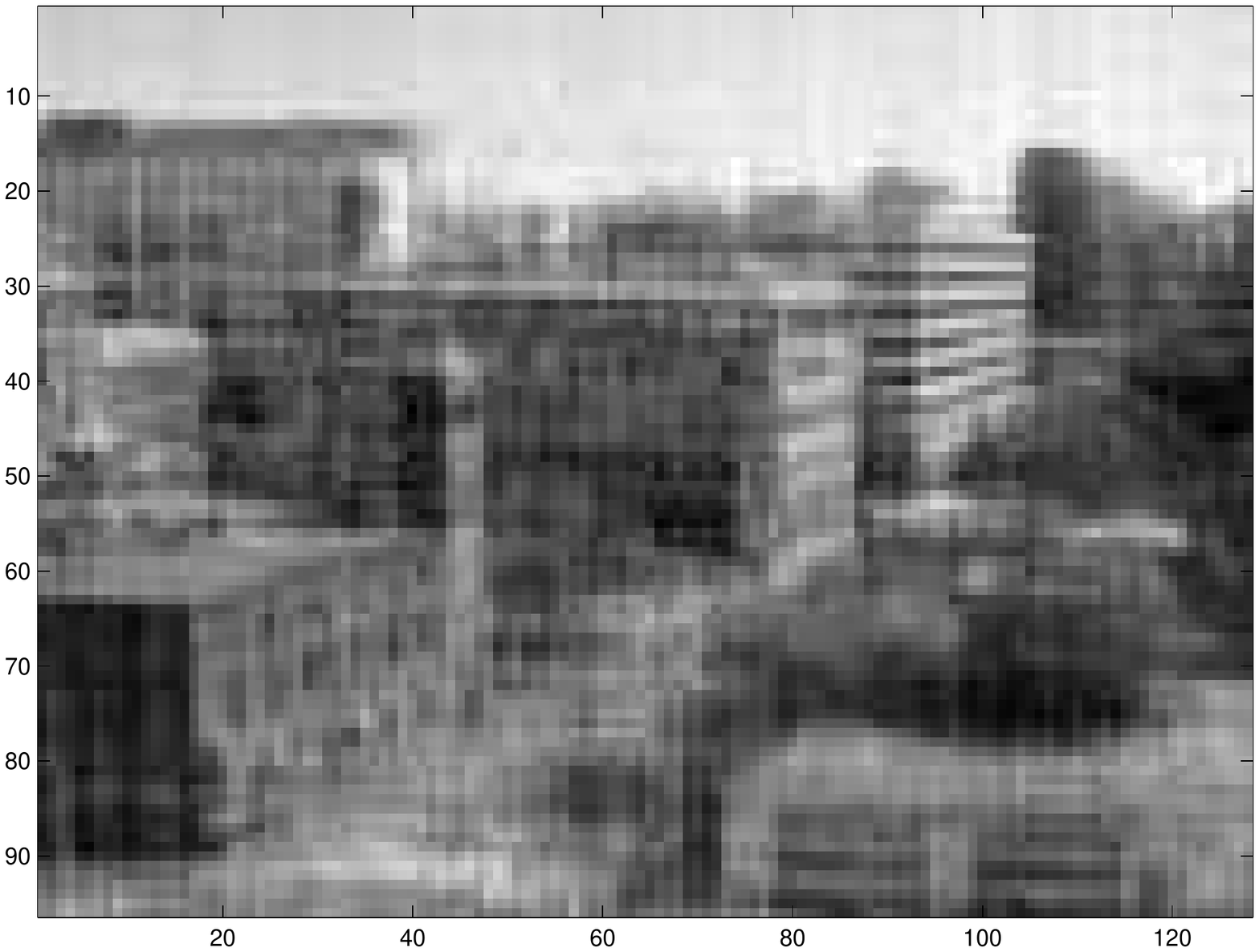}\\
        \vspace{-20mm}\\
     \includegraphics[width=0.18\textwidth]{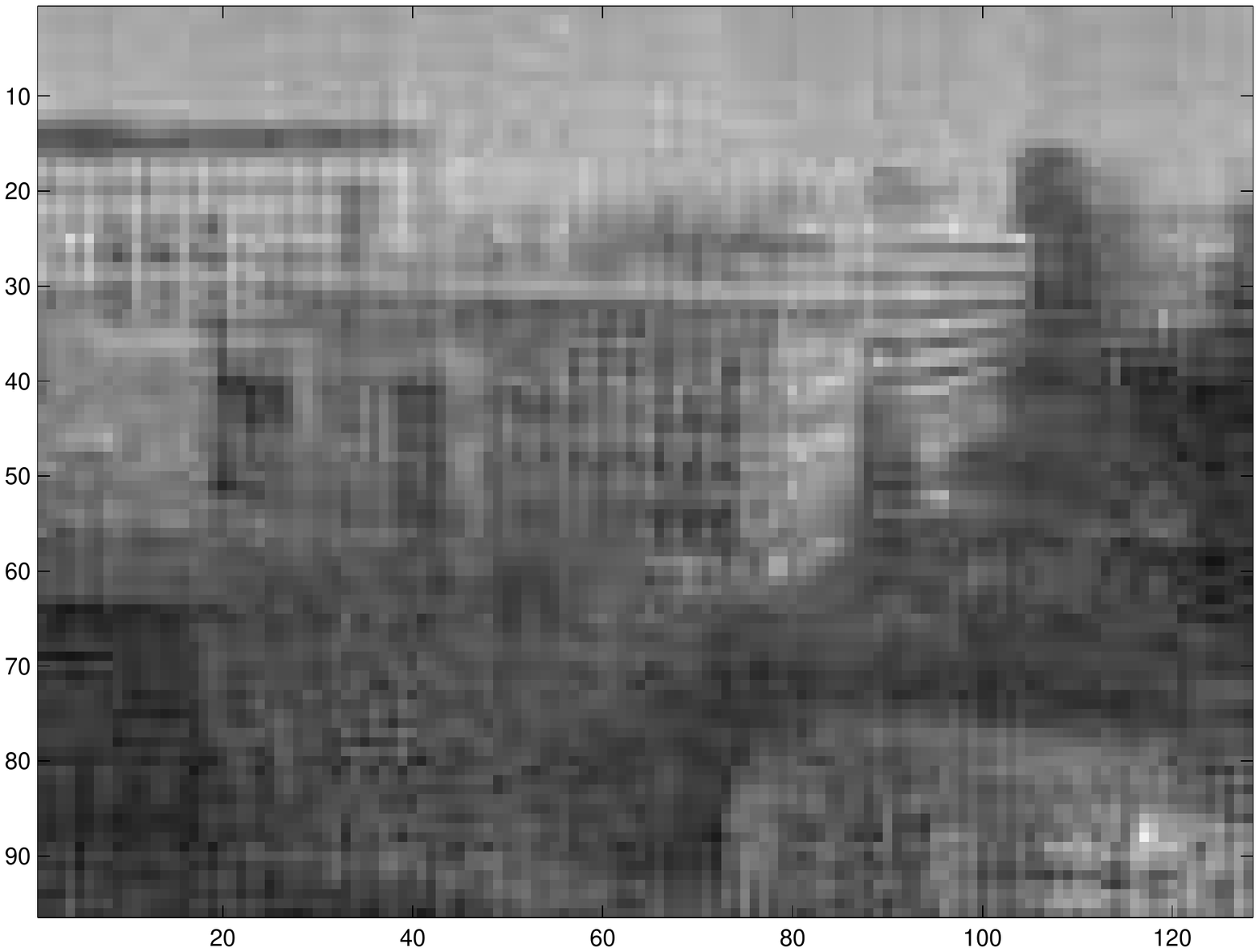} & \hspace{-5mm}  \includegraphics[width=0.18\textwidth]{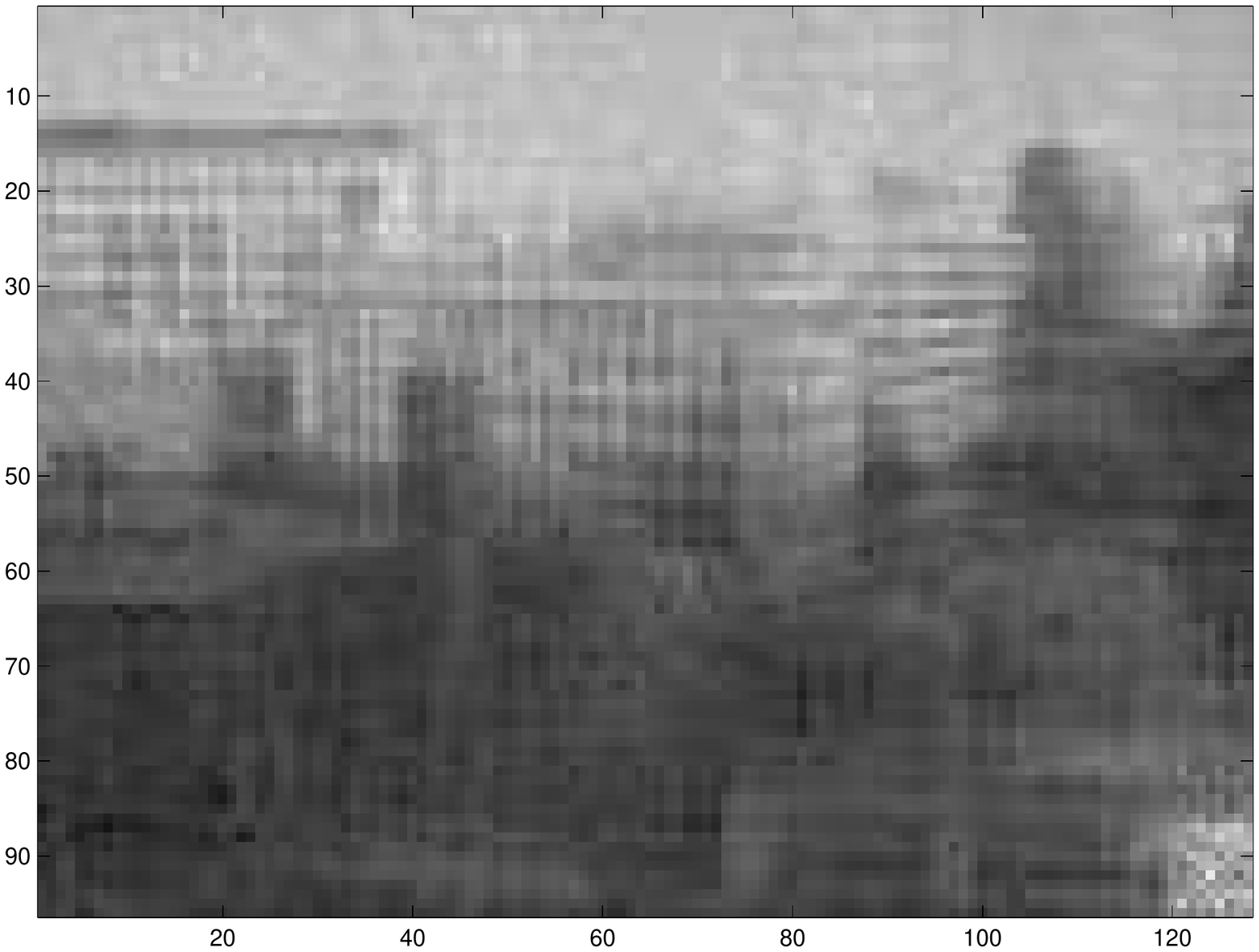}&  \hspace{-5mm} \includegraphics[width=0.18\textwidth]{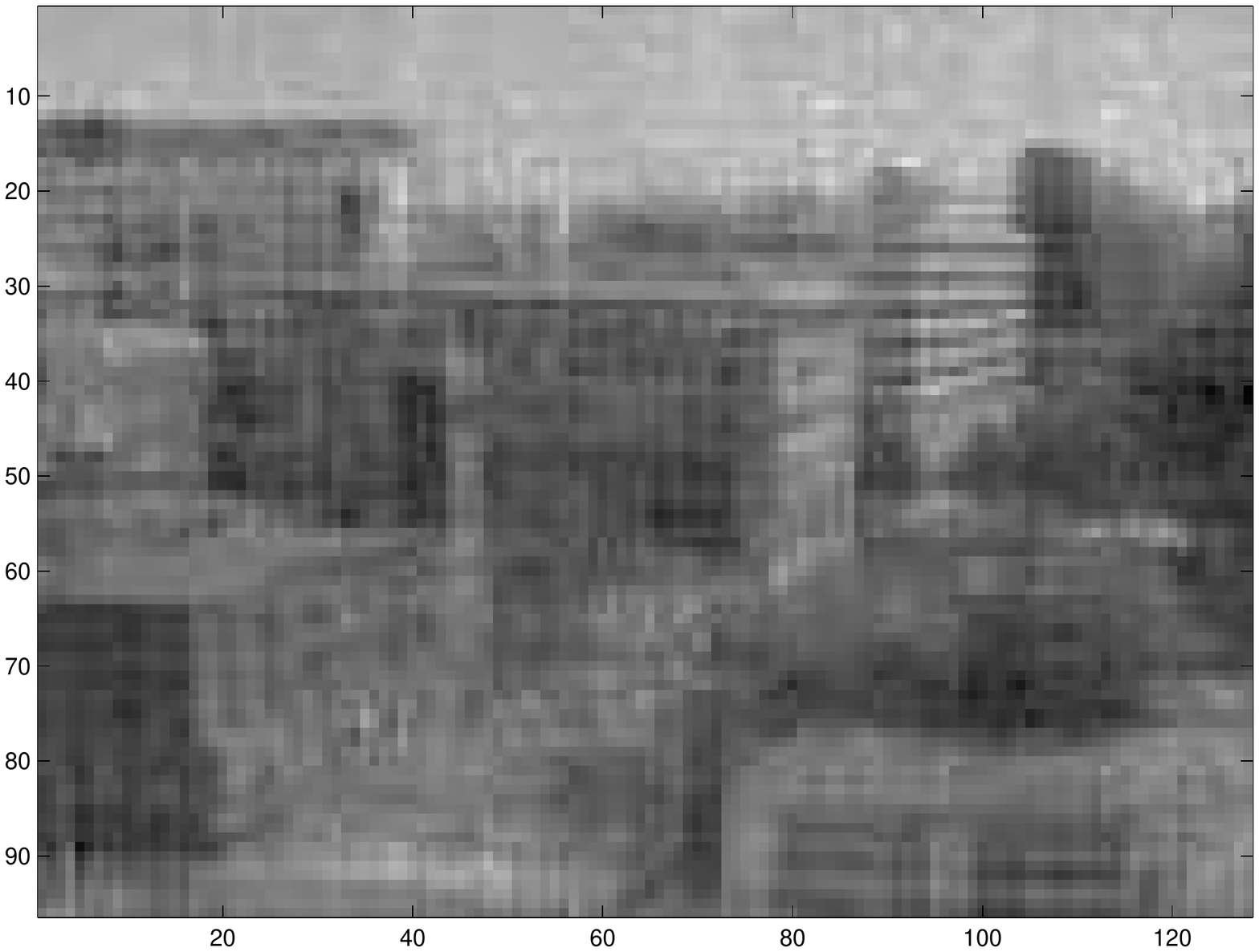}\\
    \vspace{-20mm}\\
    \includegraphics[width=0.18\textwidth]{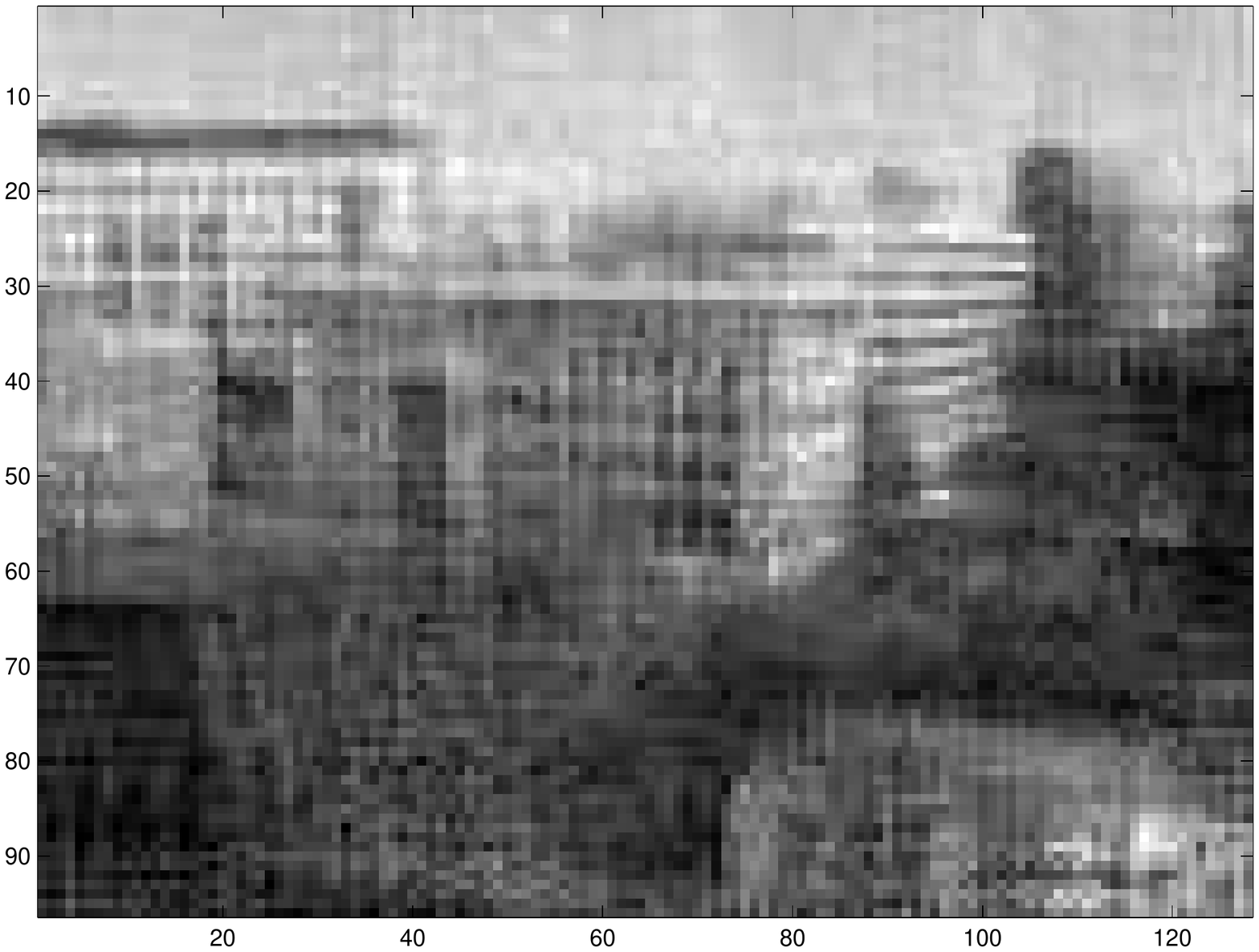} & \hspace{-5mm}  \includegraphics[width=0.18\textwidth]{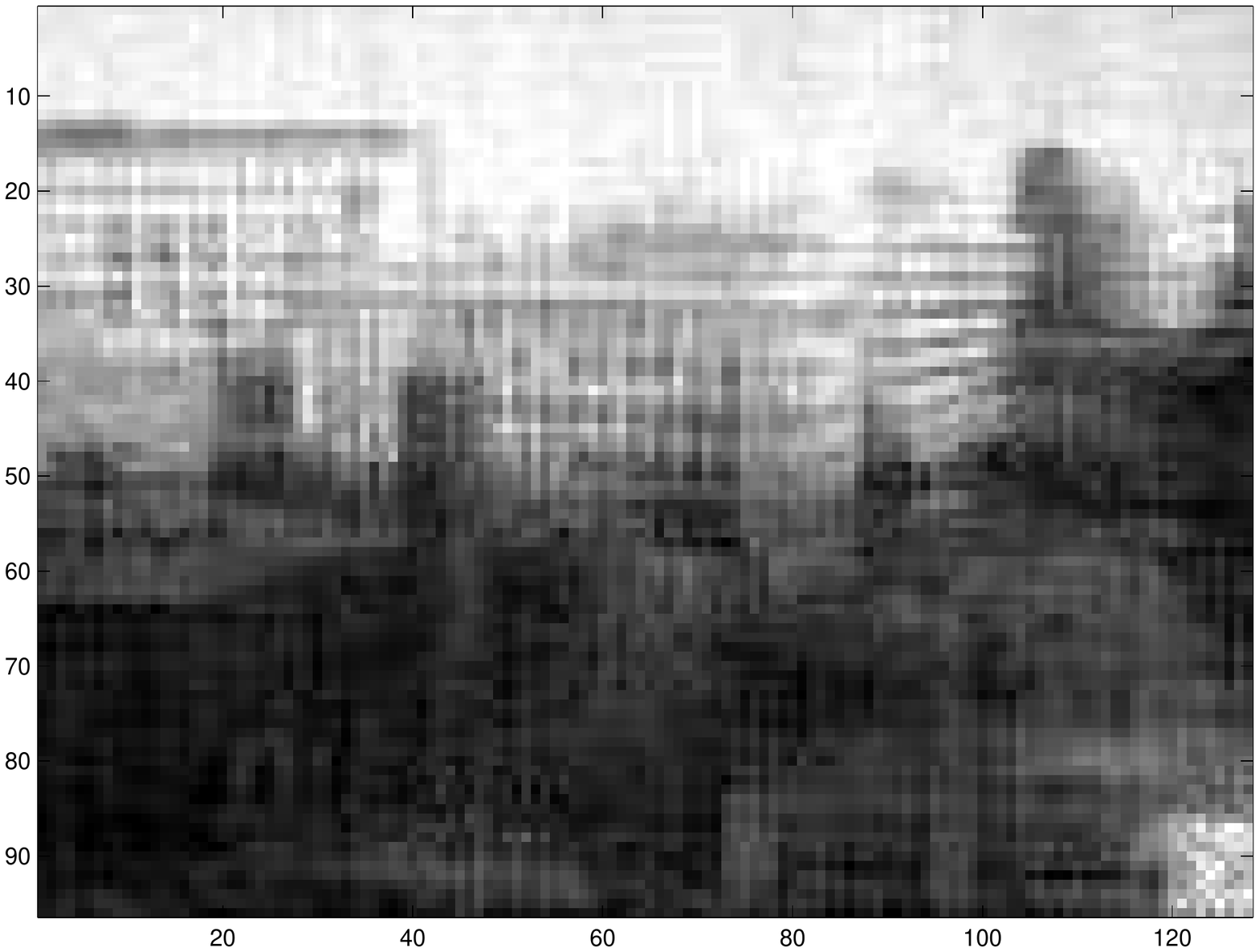}&  \hspace{-5mm} \includegraphics[width=0.18\textwidth]{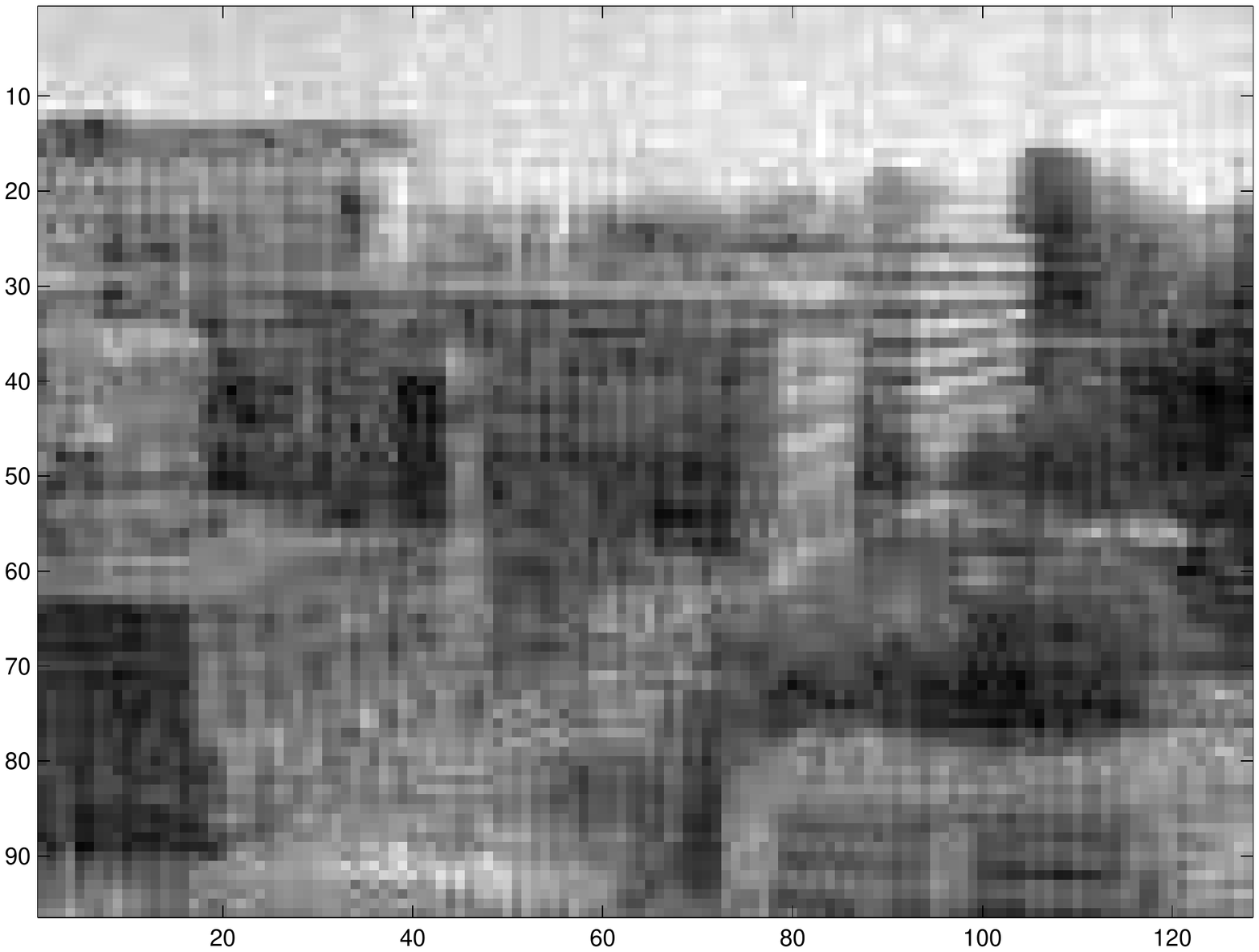}\\
    \end{tabular}}
          \vspace{-8mm}
\caption{\footnotesize{Left to right: frame 1, 15, and 25. From top row to bottom row: Original video, FoReL, OTEG, OTEG with entries truncated to $[-1,1]$. Note the visual difference in performance.}}
\label{fig:data}
\end{figure}We now experimentally demonstrate the effectiveness of OTEG on a 3-D video data. The test data, which we call $\T{M}$, is a 96x128x38 black and white video of a time-lapse city scene (Fig. \ref{fig:data}), where each pixel is a light intensity in $[-1,1]$. The reason to choose this as the working example is because it can model dynamically changing rating  matrices, and we can evaluate the performance \emph{visually}. In particular, the shadow that propagates across the city can be seen as a ``trend'' emerging and fading in a time-dynamic collaborative filtering setting. 


 We simulate online learning of this data as follows: at time $t$, a pixel from the image is sampled at random from a uniform distribution among the set of pixels which have not yet been played, and this pixel is played as $y_t$. We apply two algorithms in this setup: (1) a partial implementation of OTEG, and (2) a standard follow-the-regularized-leader approach with  tensor-nuclear-norm regularization. 
 
\begin{figure}
\centering \makebox[0in]{
    \begin{tabular}{c}
       \vspace{-28mm}\\
    \includegraphics[width=0.40\textwidth]{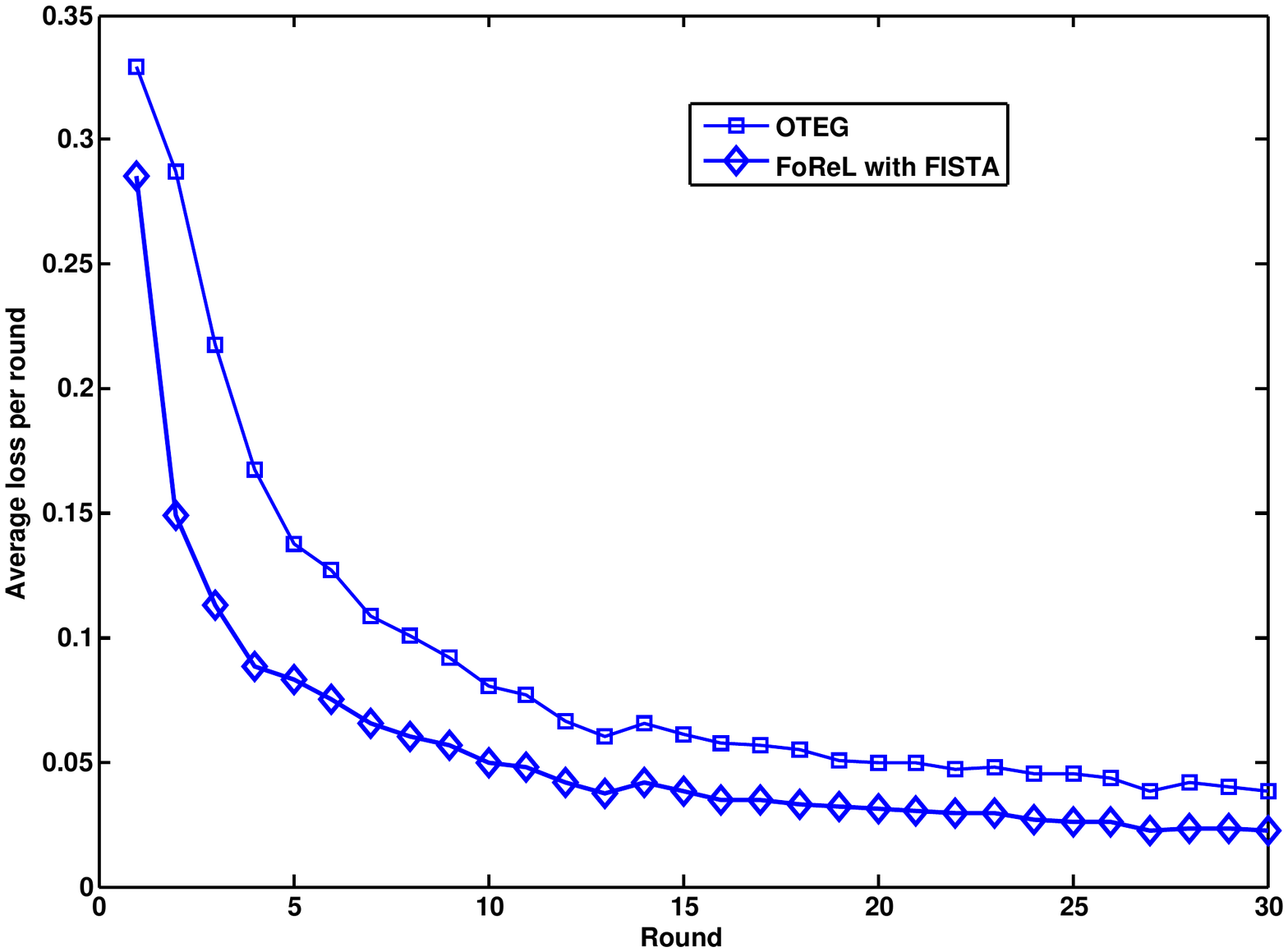}
    \end{tabular}}
    \vspace{-20mm}
\caption{\footnotesize{Loss plots for OTEG  and FoREL after 70000 iterations.}}
\label{fig:loss}
\end{figure}  For both experiments, $n=96$, $m=128$, $d=38$ , $T=70000$ (15\% of the video), and $l_t(p_t)=(y_t-p_t)^2$. The loss plots show a moving average of the value of $l_t(p_t)$, where the $T$ time intervals are divided into $R=30$ \emph{rounds} and we plot the average loss over each round.

\textbf{Discussion}--The loss plot and pictorial representation of frames 1,15, and 25 of the final iterate for both experiments are shown in Fig. \ref{fig:data} and Fig. \ref{fig:loss}. We see that in this case the FoReL algorithm achieves faster convergence, slightly better asymptotic performance, and better visual recovery of the image.  



\begin{figure}
\centering \makebox[0in]{
    \begin{tabular}{c}
       \vspace{-25mm}\\
    \includegraphics[width=0.4\textwidth]{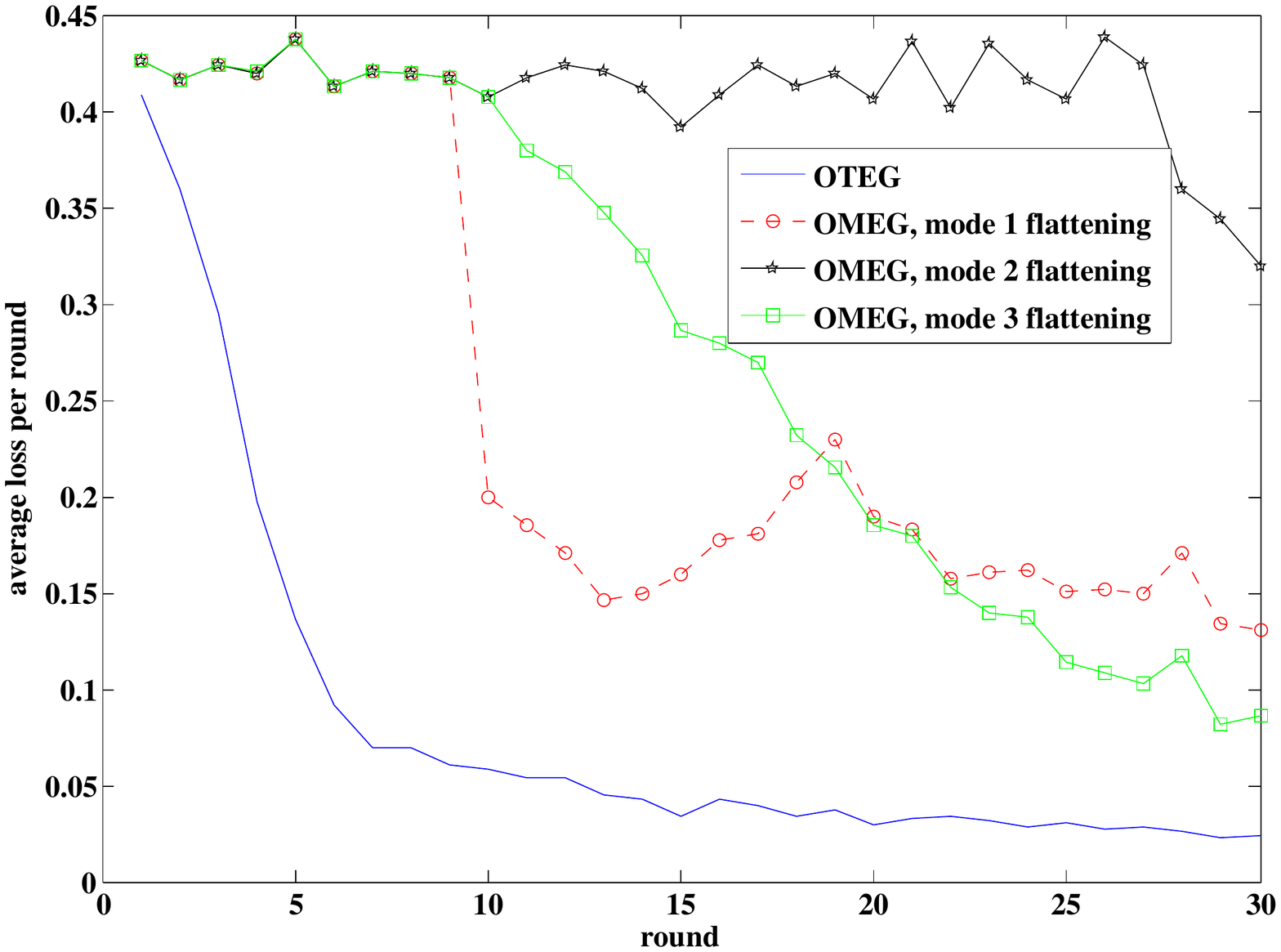}
    \end{tabular}}
    \vspace{-17mm}
\caption{\footnotesize{Loss plots for OTEG  OMEG on a reduced size data cube. Note the superior performance of OTEG compared to OMEG.}}
\label{fig:loss_OMEG}
\end{figure} 


\textbf{Comparison with Online Matrix Exponentiated Gradient (OMEG) Descent} - We again implemented a version of the OMEG algorithm, \cite{Hazan2012:un}, using OTEG but with the third dimension set to 1. The data for comparison was the same video as in the previous section but reduced to size $58\times 77 \times 20$. For OMEG, the 3-D data cube was flattened to a 2-D matrix in the 3 possible ways or \emph{modes}, \cite{TomiokaSHK11}. The total number of plays for this set-up was chosen to be $T = 17,864$ which is about $20$\% of the data. Note that the error performance of the best possible mode flattening for OMEG is much worse than the OTEG performance. 

\section{Conclusion and Future Work}
In this paper we presented an extension of strategies for online learning and prediction of matrices to tensors. Theoretical performance guarantees are derived which parallel the guarantees for the matrix case.  

We demonstrated the utility of the proposed algorithm on several test cases. In future we will extend this work to consider higher order tensors and compare the performance with methods, which use different algebraic approaches for tensor factorization.

\section{Appendix}

\begin{lemma}
\label{lem:grad}
For a real valued function $f$ with real valued domain defined via $f(\T{w}) = \hat{f}(\widehat{\T{w}})$, the gradient $\nabla_{\T{W}} f(\T{W}) = \ifft(\nabla_{\widehat{\T{W}}} \hat{f}(\widehat{\T{W}}),[\,],3).$
\end{lemma}
The proof of the Lemma follows from Lemma \ref{lem:grad1}, which is a standard result in complex analysis.

\begin{lemma}
\label{lem:grad1}
Let $g: \Cplx^n \to \Cplx^n$ be holomorphic and $f: \Cplx^n \to \Real$ be real-differentiable. Then, the complex gradient of $f \circ g$ is given by
\[
\nabla (f \circ g) = 2\frac{\partial f(g)}{\partial \bar{z}} = (\nabla f){(\frac{\partial \bar{g}}{\partial \bar{z}})}
\]
\end{lemma}
Proof is a simple consequence of the Cauchy-Riemann condition on $g$ and Equation (33) in \cite{Degaldo09}. We now prove Lemma \ref{lem:grad}.
\begin{proof}
It will be helpful to re-parameterize $f$ as a function of an arbitrary tube ${\V{w}}_{ij} = \T{W}(i,j,:)$ and the remainder of the tensor $\T{W}_{\backslash {\V{w}}_{ij}}$. We let ${\V{w}}_{ij}$ be complex, and calculate the partial complex gradient $\nabla_{{\V{w}}_{ij}} f(\T{W})$. By Lemma \ref{lem:grad1}, we have
\begin{align*}
\nabla_{{\V{w}}_{ij}} f(\T{W})
&= \nabla_{\V{w}_{ij}} f(\T{W}_{\backslash {\V{w}}_{ij}},\V{w}_{ij}) \nonumber \\
&= \nabla_{\V{w}_{ij}} \hat{f}(\widehat{\T{W}}_{\backslash {\hat{\V{w}}}_{ij}},\M{F}{{\V{w}}}_{ij}) \nonumber \\
&= \nabla_{\V{z}} \hat{f}(\widehat{ \T{W}}_{\backslash {\hat{\V{w}}}_{ij}},\V{z})(\frac{\partial}{\partial \bar{\V{z}}}\overline{\M{F}\V{z}}) \\
&= [\nabla_{\widehat{\T{W}}}\hat{f}(\widehat{\T{W}})](i,j,:) \overline{\M{F}}
\end{align*}
Where the last equation follows from the fact that $\overline{\M{F}\V{z}} = \bar{\M{F}}\bar{\V{z}}$. To see that this proves the result, recall that we chose to represent a tube as a column vector, but this orientation was arbitrary. It follows that the row-vector interpretation of a tube will give 
\begin{align*}
\nabla_{{\V{w}}_{ij}} f(\T{W}) & = ([\nabla_{\widehat{\T{W}}}\hat{f}(\widehat{\T{W}})](i,j,:) \bar{\M{F}})^\top \nonumber \\
&= \M{F}^{\dagger}([\nabla_{\widehat{\T{W}}}\hat{f}(\widehat{\T{W}})](i,j,:))^\top \\
&= \ifft([\nabla_{\widehat{\T{W}}}\hat{f}(\widehat{\T{W}})] (i,j,:))
\end{align*}
\end{proof}

\subsection{The Block-Diagonal Linear \texorpdfstring{$(\V{\beta},\V{\tau})$}{(B,t)} Game}
\label{sec:main_proof}
We consider a linear, positive-definite variation of Online Tensor Prediction, which has the same setup as in the previous Section. Let $\mcr{S} \in \{ \mcr{A} \subseteq \mcr{S}^{N \times N \times d}_{++} : \forall \T{W} \in \mcr{A},\forall k : \mathrm{Tr}(\widehat{\M{W}}^{(k)})\leq\tau(k),\forall k\forall i: \widehat{\M{W}}^{(k)}(i,i)\leq \beta(k)\}$ be a set of positive-definite tensors with Fourier domain trace and diagonal-entry bounds $\V{\tau}$ and $\V{\beta}$, respectively. The loss functions $l_t$ will have the form $l_t(\T{W}_t) = \mathrm{Tr}(\T{W} \star\T{L}_t) = \mathrm{Tr}(\blkd( \widehat{\T{W}})\blkd(\widehat{\T{L}}_t))$ for tensors $\T{L}_t$. This problem is nearly identical to the linear game discussed in \cite{Hazan2012:un}, except now every matrix is block-diagonal, and there is an independent $\beta$ and $\tau$ constraint for each block. Thus, we can apply the general regret bound derived for the original game, with slight modification. As required by the proof, we assume that the spectral norm $||\eta\blkd(\widehat{\T{L}}_t)|| \leq 1$ for all $t$.

\begin{theorem}
\label{thm:TEG_regret2}
Suppose Tensor Exponentiated Gradient is run on Linear Online Tensor Prediction. Then, 
\[
\mathrm{Regret} \leq \sum_{k=1}^d\{ \eta\sum_{t=1}^T \mathrm{Tr}(\widehat{\M{W}}_t^{(k)}(\widehat{\M{L}}_t^{(k)})^2) + \frac{\tau(k)\log{N}}{\eta}\}
\]
\end{theorem}

The proof follows along the same lines as that of Theorem 10 in \cite{Hazan2012:un}, except for the fact that we split trace operations by summing over the traces of each face. The fact that our matrices are complex does not matter, as the given proof is valid for all Hermitian matrices. Specifically, both the Golden Thompson inequality and the relation $\exp(\M{A}) \preceq \M{I} + \M{A} + \M{A}^2$ hold for general Hermitian $\M{A}$ with spectral norm $||\M{A}|| \leq 1$. Note that for this result to be meaningful, we need to bound $\mathrm{Tr}((\widehat{\M{L}}_t^{(k)})^2)$. Let us further assume, then, that $\mathrm{Tr}((\widehat{\M{L}}_t^{(k)})^2) \leq \gamma(k)$ for some $\V{\gamma} \in \mathbb{R}^d$. For this scenario, Theorem~\ref{thm:TEG_regret2} gives the following corollary.

\begin{corollary}
\label{cor:Cor1}
Suppose TEG is run on $\V{\gamma}$-constrained Linear Online Tensor Prediction. Then,
\begin{align*}
\mathrm{Regret} &\leq \sum_{k=1}^d\{\eta T\beta(k)\gamma(k)  + \frac{\tau(k)\log{N}}{\eta}\} \notag \\
& = \eta T \sum_{k=1}^d\beta(k)\gamma(k) + \frac{\log N}{\eta} \sum_{k=1}^d \tau(k)
\end{align*}
\end{corollary}
With this result in hand, we have a guide to deriving a bound for the general game, where loss functions are not necessarily linear.

\subsection{Proof of Theorem~\ref{thm:TEG_regret}}
We first analyze Algorithm \ref{alg:OTEG}. For this we find a linear approximation of the regret, which will permit us to use Corollary \ref{cor:Cor1}. For any $\T{U} \in \mcr{S}$, note that
\begin{align*}
&\mathrm{Tr}(\blkd(\widehat{\phi}(\T{U}))\blkd(\widehat{\T{L}}_t))  \notag \\
&= g\sum_{k=1}^d  (\widehat{\M{P}}^{(k_t)}(i_t,j_t)-\widehat{\M{N}}^{(k_t)}(i_t,j_t))\overline{\M{F}}(k,k_t) \\
& \hspace{10mm} +  (\widehat{\M{P}}^{(k_t)}(j_t,i_t)-\widehat{\M{N}}^{(k_t)}(j_t,i_t))\M{F}(k,k_t) \\
& = \frac{g}{\sqrt{d}}\sum_{k=1}^d  \widehat{\M{U}}^{(k_t)}(i_t,j_t)e^\frac{2\pi i (k-1)(k_t-1)}{d} \notag 
\\ & \hspace{12mm}+ \widehat{\M{U}}^{(k_t)\dagger}(i_t,j_t) e^\frac{-2\pi i (k-1)(k_t-1)}{d} \\
&= 2g \ \ifft(\widehat{\T{U}})(i_t,j_t,k_t) = 2g\T{U}(i_t,j_t,k_t)
 \end{align*}
We see that the linear loss $\mathrm{Tr}(\blkd(\widehat{\T{W}})\blkd(\widehat{\T{L}}_t))$ is equivalent to $2gP_{\widehat{\T{W}}}(i_t,j_t,k_t)$, and thus
\begin{align*}
\mathrm{Tr}(\blkd(\widehat{\T{W}})\blkd(\widehat{\T{L}}_t)) & = 2gP_{\widehat{\T{W}}_t}(i_t,j_t,k_t) = 2gp_t
\end{align*}
This implies that, 
\begin{align*}
& \mathrm{Tr}(\blkd(\widehat{\T{W}})\blkd(\widehat{\T{L}}_t)) - \mathrm{Tr}(\blkd(\widehat{\phi}(\T{U}))\blkd(\widehat{\T{L}}_t)) & \nonumber \\
& = 2g(p_t - \T{U}(i_t,j_t,k_t)) \geq 2(l_t(p_t) - l_t(\T{U}(i_t,j_t,k_t)))
\end{align*}
By convexity of $l_t$. Thus, the regret of our algorithm is at most half the regret of the linear game with loss tensors $\widehat{\T{L}}_t$.

Assuming $\eta ||\blkd(\widehat{\T{L}}_t)|| \leq 1$, we apply Corollary \ref{cor:Cor1} to obtain
\begin{align*}
\mathrm{Regret} & \leq \frac{1}{2}\mathrm{Regret}_{\mathrm{Linear}} \\
& \leq \frac{1}{2}\left\{4G^2 \eta T \sum_{k=1}^d \beta(k) + \frac{\log N}{
\eta}\sum_{k=1}^d \tau(k)\right\}
\end{align*}
Setting $\eta = \sqrt{\frac{\log N \sum_{k=1}^d \tau(k)}{4G^2 T\sum_{k=1}^d \beta(k)}}$, we have
\begin{align*}
\mathrm{Regret} &\leq \frac{1}{2}\left\{2\sqrt{4G^2T\log N (\sum_{k=1}^d \beta(k))(\sum_{k=1}^d \tau(k))}\right\} \\
&= 2G\sqrt{T\log N (\sum_{k=1}^d \beta(k))(\sum_{k=1}^d \tau(k))}
\end{align*}
Which gives us the stated bound. 

We now address the technical condition that $\eta ||\blkd(\widehat{\T{L}}_t)|| \leq 1$. Let $k_0 = \arg\max_{k \in [d]}||\widehat{\M{L}}^{(k)}_t||$. We have $||\blkd(\widehat{\T{L}}_t)|| = \max_{k \in [d]}||\widehat{\M{L}}^{(k)}_t|| = ||\widehat{\M{L}}_t^{(k_0)}|| \leq ||\widehat{\M{L}}_t^{(k_0)}||_F = \sqrt{\mathrm{Tr}((\widehat{\M{L}}_t^{(k_0)})^2)} \leq \sqrt{\gamma(k_0)} = \sqrt{4G^2}$. For $T \geq \frac{\log N \sum_{k=1}^d \tau(k)}{\sum_{k=1}^d \beta(k)}$ and our choice of $\eta$, this implies $\eta ||\blkd(\widehat{\M{L}}_t)|| \leq 1$, and the bound holds. 

For $T < \frac{\log N \sum_{k=1}^d \tau(k)}{\sum_{k=1}^d \beta(k)}$, note that $l_t$ have derivatives bounded by $G$, and the domain is $[-1,1]$, so the maximum possible regret on any round is $2G$. Hence the regret up to time $T$ is at most $2GT < 2G\sqrt{T\log N (\sum_{k=1}^d \beta(k))(\sum_{k=1}^d \tau(k))}$, since $||\beta||_1 \geq 1$.

\bibliographystyle{IEEEbib}
\bibliography{OPM.bib}

\end{document}